\documentclass[11pt]{article}

\usepackage[
  twocolumnmode,              
  shownumpages,               
  bgcolor={245,245,250},      
  braincolor={190,32,240},    
  citingstyle=authoryear,     
  bibliostyle=plainnat,       
  bibfile=refs          
]{brainlab}

\usepackage{amsthm}

\theoremstyle{plain}
\newtheorem{proposition}{Proposition}
\newtheorem{corollary}{Corollary}

\setbrainmeta{
  title={Why SGD is not Brownian Motion: A New Perspective on Stochastic Dynamics},
  authors={
      Igor Ignashin\textsuperscript{1},
      Anna Radovskaya\textsuperscript{1,3},
      Andrew Semenov\textsuperscript{1,3},
      Egor Lopatin\textsuperscript{1},
      Stanislav Olegovich Potapov\textsuperscript{1},
      Aleksandr Kovalenko\textsuperscript{1,3},
      Andrey Veprikov\textsuperscript{1},
      Aleksandr Shestakov\textsuperscript{1},
      Andrey Leonidov\textsuperscript{1,3},
      Aleksandr Beznosikov\textsuperscript{1,2}
    },
    affiliations={
      \textsuperscript{1}Basic Research of Artificial Intelligence Laboratory (BRAIn Lab) \\
      \textsuperscript{2}Innopolis University \\
      \textsuperscript{3}P.N. Lebedev Physical Institute of the Russian Academy of Sciences
    },
  abstract={
  Stochastic Gradient Descent (SGD) is commonly modeled as a Langevin process, assuming that minibatch noise acts as Brownian motion. However, this approximation relies on a continuous-time limit and a $\sqrt{\eta}$ noise scaling that does not match the discrete SGD update at finite learning rate.
    In this work, we propose an alternative formulation of SGD as deterministic dynamics in a fluctuating loss landscape induced by minibatch sampling. Starting directly from the discrete update, we derive a master equation for the parameter distribution and obtain a discrete Fokker–Planck equation that differs from the standard Langevin form at order $\eta^2$.    
    Using this framework, we analyze SGD dynamics near critical points of the loss. We show that the behavior decomposes along the eigenbasis of the mean Hessian into qualitatively distinct regimes. In particular, nearly-flat directions do not admit a stationary distribution: the variance grows over time, corresponding to effective diffusion along valleys with a coefficient proportional to the learning rate.
    We provide empirical evidence supporting these predictions on neural network models in computer vision and natural language processing, observing a clear qualitative separation between confined and diffusive modes.
    }
}

\begin{document}
\begin{mainpart}

\section{Introduction}

Stochastic Gradient Descent (SGD) has been the primary optimization method for training neural networks since the early days of deep learning \citep{rumelhart1986learning}. Despite significant advances in model architectures, datasets, and computational scale, SGD remains a central component of modern machine learning pipelines \citep{bottou2010large, goodfellow2016deep}. It is widely used to train over-parameterized models such as Transformers \citep{vaswani2017attention} and large-scale language models \citep{team2025kimi, hernandez2025apertus}, and serves as the foundation for many adaptive optimization methods, including Adam \citep{kingma2014adam} and more recent variants such as Muon \citep{jordan2024muon}. In fact, recent empirical studies \citep{sreckovic2025your} indicate that for sufficiently small batch sizes, the performance of adaptive optimizers approaches that of vanilla SGD, reinforcing its role as a fundamental baseline. These observations highlight that, despite its simplicity, SGD remains both practically indispensable and theoretically important.

Despite its widespread use, a complete theoretical understanding of SGD dynamics is still lacking. A dominant line of work interprets SGD through the lens of statistical physics, modeling it as a noisy relaxation process similar to Langevin dynamics \citep{mandt2017stochastic, simsekli2019tail, xie2021a}. In this view, stochastic gradients are treated as Gaussian perturbations around the gradient of the average loss, leading—under simplifying assumptions—to a stationary Gibbs distribution. While this framework provides useful intuition, it relies on assumptions such as Gaussian gradient noise and the existence of a stationary regime, which may not hold in practical deep learning settings. As discussed in Section~\ref{sec:rw}, empirical and theoretical studies increasingly point to systematic deviations from this Brownian-motion-based picture.

An important step toward understanding these deviations was made by \citet{feng2021inverse}, who established an inverse variance–flatness relation for SGD near solutions. Their results show that directions with smaller curvature exhibit smaller fluctuations, revealing a nontrivial connection between loss geometry and stochastic dynamics. However, this relation is derived under specific assumptions and does not provide a general dynamical framework for SGD.

In this work, we revisit the foundations of SGD dynamics and show that, at finite learning rates, the standard Langevin approximation does not faithfully reproduce the discrete SGD dynamics. Rather than modeling SGD as Brownian motion in a fixed potential, we interpret it as deterministic dynamics in a \emph{fluctuating loss landscape}, where randomness arises from minibatch sampling. Unlike physical systems, where noise is external and uncontrollable, SGD stochasticity is algorithmically generated and directly tied to the structure of the loss function. This distinction allows us to derive the stochastic dynamics from first principles, without relying on continuous-time approximations or phenomenological noise models. Importantly, our results do not contradict continuous-time diffusion approximations in the infinitesimal step-size limit, but instead highlight discrepancies that arise at finite step size.

\paragraph{Contributions.}
Our main contributions are as follows:
\begin{itemize}
    \item \textbf{Discrete Fokker--Planck framework.}  
    Starting from the exact SGD update, we derive a discrete Fokker--Planck equation governing the evolution of the parameter distribution. We show that standard Langevin-based approximations omit terms of order $\eta^2$, including contributions of the same order as the retained diffusion term, which can lead to qualitatively incorrect predictions at finite learning rates.
    
    \item \textbf{Analysis near critical points.}  
    We analyze SGD dynamics in the vicinity of critical points of the loss function using a quadratic approximation. This yields explicit expressions for the variance of parameter trajectories in terms of Hessian statistics.
    
    \item \textbf{Unified picture of SGD dynamics.}  
    Our framework reveals distinct dynamical regimes in the Hessian eigenbasis. Directions with positive curvature exhibit bounded fluctuations consistent with the inverse variance–flatness relation \citep{feng2021inverse}, while nearly-flat directions remain \ non-stationary and exhibit diffusive behavior.
    
    \item \textbf{Empirical validation.}  
    We validate these predictions on vision and language models by analyzing Hessian spectra and parameter variance along eigendirections. The observed dynamics are consistent with the theoretical predictions.
\end{itemize}

\section{Related Work}
\label{sec:rw}

\paragraph{Langevin-based views of SGD.}  
A classical line of research models SGD as a discretization of Langevin dynamics, establishing connections to equilibrium statistical physics. In \citet{mandt2017stochastic}, SGD was interpreted as approximate posterior sampling near local minima, with minibatch noise acting as an effective temperature. Related work \citep{smith2017bayesian} analyzed the interplay between batch size, learning rate, and noise scale, leading to the ``noise-scale'' rule. Within this framework, SGD has been associated with implicit regularization and a preference for flat minima \citep{hu2017diffusion, chaudhari2018stochastic, zhu2019aniso, yang2023}.  

While this perspective has been influential, it relies on approximations such as Gaussian gradient noise and the existence of a stationary regime. These assumptions are not always justified in modern deep learning settings and can lead to discrepancies between the Langevin approximation and the actual discrete SGD dynamics.

\paragraph{Heavy-tailed and non-Gaussian noise.}  
Several works show that minibatch-gradient noise can be non-Gaussian or heavy-tailed, leading to dynamics that differ from Brownian motion \citep{simsekli2019tail, simsekli2020fractional, gurbuzbalaban2021heavy, nguyen2019first}. These works typically replace Gaussian noise with alternative stochastic models, whereas we derive the dynamics directly from the discrete SGD update.

\paragraph{Anisotropy and structure of gradient noise.}  
Another line of work emphasizes that minibatch noise is highly anisotropic. In \citet{jastrzkebski2018relation}, the covariance of gradient noise was shown to align with the Hessian, suggesting that the interaction between noise and curvature plays a key role in determining SGD behavior. Further work \citep{zhu2019aniso} demonstrated that anisotropic noise facilitates escape from sharp minima and biases optimization toward flatter regions.

These findings indicate that SGD noise cannot be accurately described as isotropic Brownian forcing. Instead, its structure is closely tied to the geometry of the loss landscape.

\paragraph{Alternative dynamical formulations.}  
Beyond Langevin-based models, several works have explored alternative descriptions of SGD dynamics. In \citet{yaida2020non}, finite-width neural networks were shown to exhibit non-Gaussian fluctuations, providing a different perspective on stochastic training dynamics. In \citet{feng2021phases}, SGD was analyzed in terms of dynamical phases, depending on noise and data properties. Closely related to our work, \citet{feng2021inverse} established an inverse variance–flatness relation, linking parameter fluctuations to local curvature.

These approaches emphasize that SGD operates in a non-equilibrium regime that is not fully captured by classical Langevin formulations.

\paragraph{Positioning of this work.}  
Our work differs from these lines of research in two key aspects. First, we derive the evolution of the parameter distribution directly from the discrete SGD update via a master equation, rather than starting from a continuous-time stochastic differential equation. Second, we interpret SGD noise as arising from fluctuations of the loss landscape itself, rather than as an external stochastic forcing term. This leads to a discrete Fokker--Planck description that captures finite-step effects and provides a unified explanation of the dynamical regimes observed in practice.

\section{Setup and Problem Formulation}

\paragraph{SGD with replacement.}
We consider stochastic gradient descent (SGD) with sampling with replacement, so that at each iteration the minibatch is drawn independently from the dataset. The parameter update is
\begin{equation}
	w_{n+1} = w_n - \eta \nabla L_n(w_n),
	\label{step}
\end{equation}
where $w_n$ denotes the parameters at step $n$, $L_n$ is the minibatch loss, and $\eta$ is the learning rate.

\paragraph{Brownian motion versus a fluctuating landscape.}
A common intuition is to view SGD as Brownian motion in a fixed potential, that is, as a particle driven by random forces while evolving in the average loss $\bar L$. This viewpoint leads to a Langevin-type update
\begin{equation}
 	w_{n+1} = w_n - \eta \nabla \bar L(w_n) - \sqrt{\eta}\,\xi_n,
\end{equation}
where $\xi_n$ is interpreted as a stochastic force. The $\sqrt{\eta}$ scaling is introduced to obtain a nontrivial continuous-time limit and the corresponding Fokker--Planck equation.

However, this construction does not exactly match SGD. In the true update~\eqref{step}, stochasticity enters through the minibatch loss itself and therefore appears at order $\eta$, not $\sqrt{\eta}$. As a result, the Langevin approximation does not reproduce the discrete SGD process exactly and omits terms that are relevant at finite step size (see Appendix~\ref{app:Lang}).

A more faithful interpretation is that SGD describes deterministic motion in a \emph{fluctuating loss landscape}: the objective changes from step to step because the sampled minibatch changes. In this view, the randomness is not an external forcing term but an intrinsic consequence of minibatch sampling. Figure~\ref{fig:landscape} illustrates the distinction.

\begin{figure}[t]
\centering
 	\includegraphics[width=0.45\textwidth]{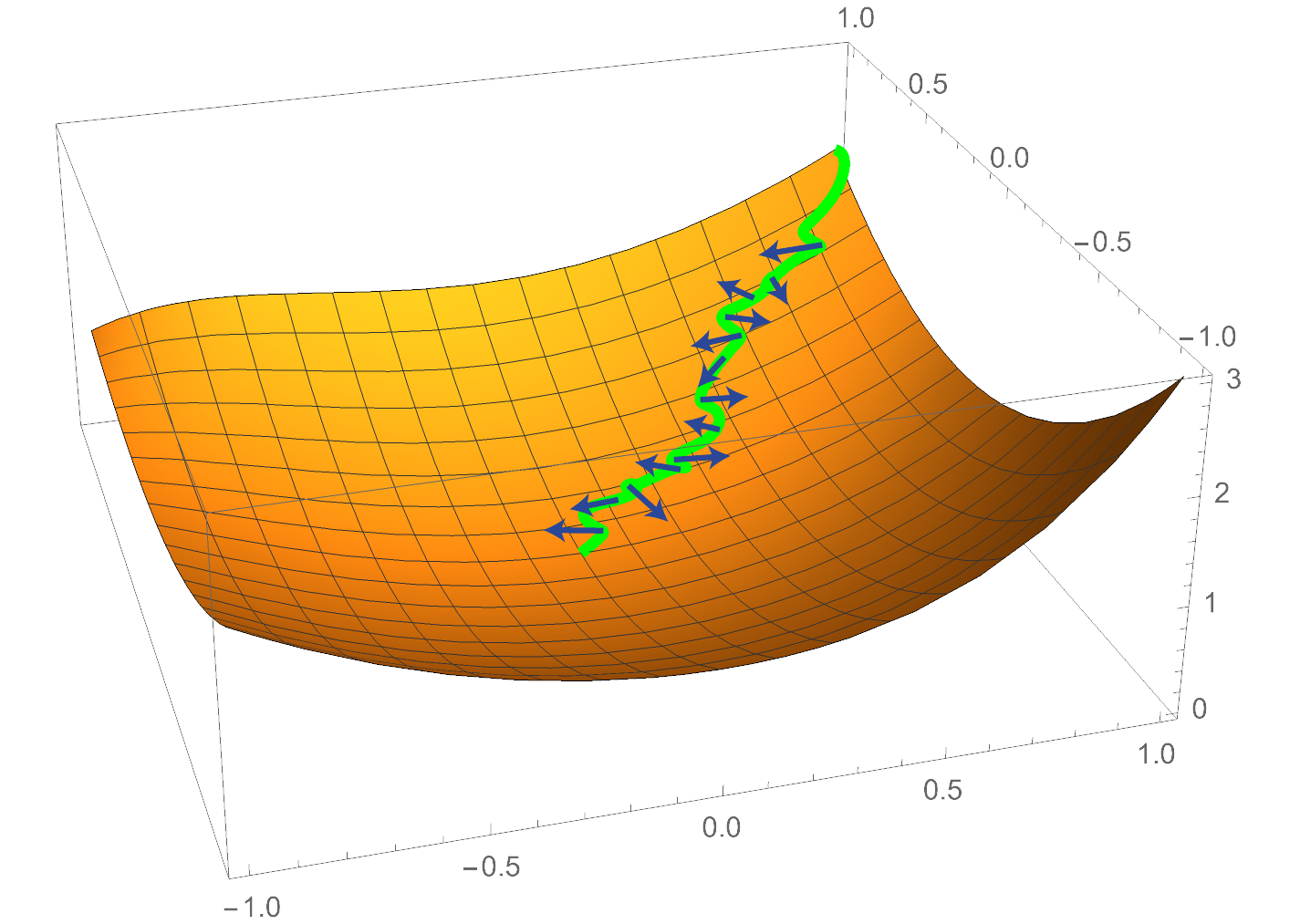}	
 	\includegraphics[width=0.45\textwidth]{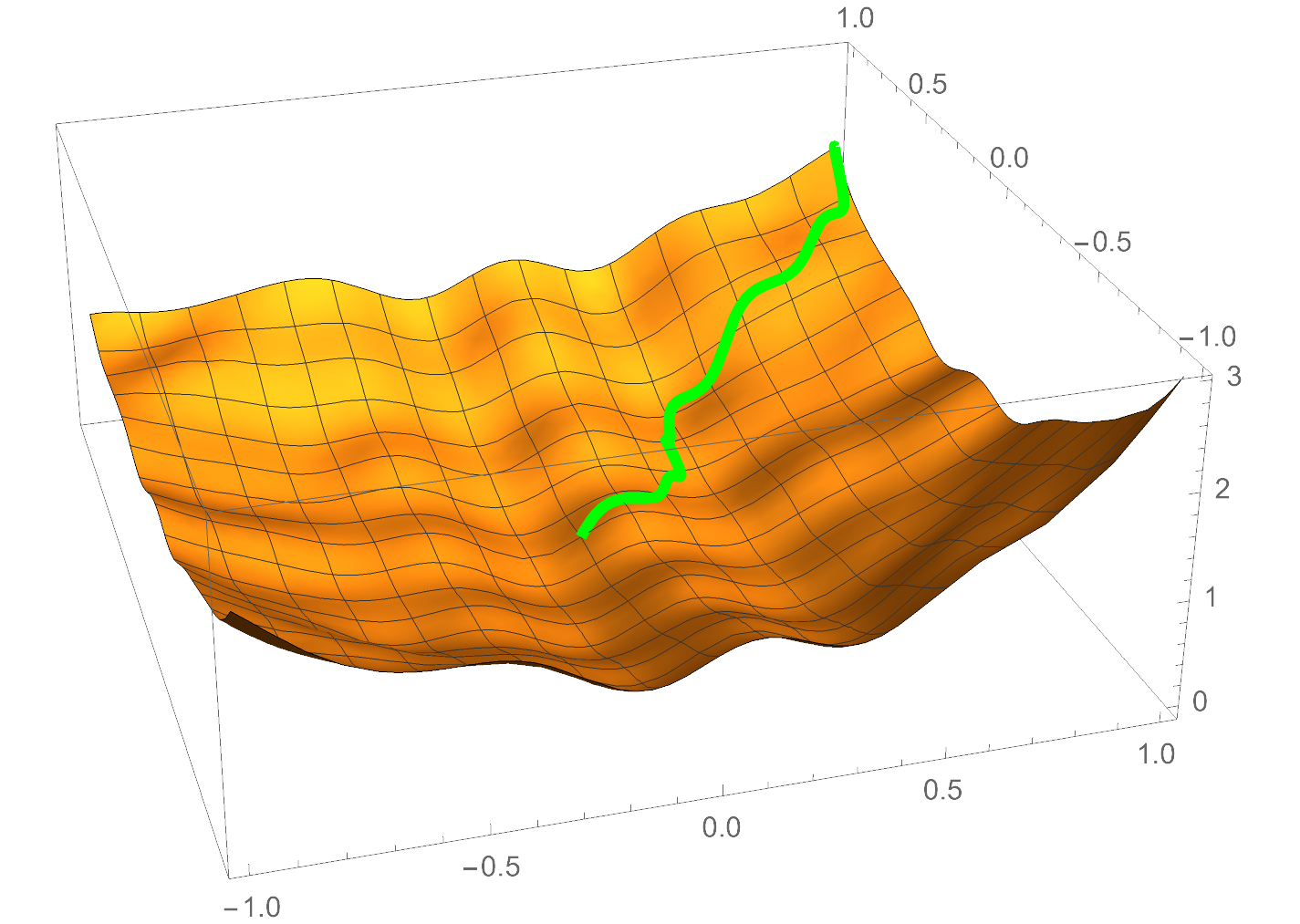}
 	\caption{Left: Brownian motion, modeled as a particle driven by random forces in a static potential. Right: deterministic motion in a fluctuating potential, which better reflects SGD with minibatch sampling.}
 	\label{fig:landscape}
\end{figure}

\paragraph{Probabilistic description.}
To study the collective dynamics of SGD, we introduce the parameter distribution $p_n(w)$ at iteration $n$. Its evolution is exactly described by the master equation
\begin{equation}
p_{n+1}(w) 
= \int_{\mathbb{R}^d} p_n(v)\;
\mathbb{E}_L\Big[
\delta\big(w - v + \eta \nabla L(v)\big)
\Big]\, dv,
\label{P}
\end{equation}
where $\delta(\cdot)$ is the Dirac delta function and $\mathbb{E}_L[\cdot]$ denotes expectation over minibatch sampling.

Equation~\eqref{P} is exact for SGD with replacement because minibatches are sampled independently across iterations. Without replacement, consecutive updates become correlated, and the corresponding master equation is more involved.

\section{Main Theoretical Results and Analysis}

\paragraph{Discrete Fokker--Planck equation.}
\begin{proposition}[Discrete Fokker--Planck equation for SGD]
\label{prop:discrete-fp}
Assume that minibatches are sampled independently with replacement, the minibatch loss
$L(w)$ is sufficiently smooth, and the required gradient moments are finite. Then the
SGD update
\[
w_{n+1}=w_n-\eta \nabla L_n(w_n)
\]
induces the following second-order expansion of the parameter density:
\begin{equation}
\label{eq:FP}
\begin{aligned}
p_{n+1}(w)
&= p_n(w)
 + \eta \sum_{k=1}^d
 \nabla_k \Big(
 p_n(w)\,\nabla_k \bar L(w)
 \Big)
\\
&\quad
 + \frac{\eta^2}{2}
 \sum_{k,l=1}^d
 \nabla_k \nabla_l
 \Big(
 \mathbb{E}_L[
 \nabla_k L(w)\nabla_l L(w)
 ]\,p_n(w)
 \Big)
\\
&\quad
 + \mathcal{O}(\eta^3).
\end{aligned}
\end{equation}
\end{proposition}

Proposition~\ref{prop:discrete-fp} is proved in Appendix~\ref{app:Lang}, where we derive the result by expanding the exact master equation for the SGD transition kernel and truncating the resulting series at second order in $\eta$. Equation~\eqref{eq:FP} is the central theoretical object of this work. It describes the evolution of the parameter distribution under SGD directly in discrete time, without imposing a continuous-time approximation.

\paragraph{Implications of Eq.~\eqref{eq:FP}.}
Several important conclusions follow immediately.

\begin{itemize}
    \item \textbf{Discrete nature of the dynamics.}  
    The equation is intrinsically discrete. In the strict limit $\eta \to 0$, the stochastic contribution disappears and SGD reduces to gradient descent.
    
    \item \textbf{Difference from Langevin scaling.}  
    Langevin approaches impose $\sqrt{\eta}$ scaling of the noise in order to preserve diffusion in continuous time. In discrete SGD, the expansion is organized in powers of $\eta$, and this changes which terms appear at a given order.
    
    \item \textbf{Role of cumulants.}  
    At order $\mathcal{O}(\eta^2)$, only the second moment of minibatch gradients enters. Higher-order cumulants appear only at higher orders.
    
    \item \textbf{Gaussian minibatch fluctuations do not imply Gaussian parameter dynamics.}  
    Even if minibatch fluctuations are approximately Gaussian, the induced parameter dynamics need not be Gaussian.
\end{itemize}

\paragraph{Mismatch with Langevin dynamics.}
Although Eq.~\eqref{eq:FP} is exact up to order $\eta^2$, it differs from the standard Langevin formulation already at this order. We now make this discrepancy explicit in a one-dimensional setting.

We consider the one-dimensional SGD update
\begin{equation}
    \theta_{n+1}=\theta_n-\eta \nabla L_n(\theta_n),
    \label{E1}
\end{equation}
with moment assumptions
\begin{equation}
\label{E2}
\begin{aligned}
\mathbb{E}_L[\nabla L_n(\theta)]
&= \nabla \bar L(\theta), \\
\mathbb{E}_L[(\nabla L_n(\theta))^2]
&= D(\theta) + \big(\nabla \bar L(\theta)\big)^2 .
\end{aligned}
\end{equation}


\begin{proposition}[Finite-step mismatch with Langevin dynamics]
\label{prop:langevin-mismatch}
Under the update~\eqref{E1} and moment assumptions~\eqref{E2}, the discrete SGD dynamics differ from the Langevin-based Fokker--Planck equation at order $\eta^2$. In particular, the discrete expansion contains an additional second-order term proportional to $(\nabla \bar L)^2$, which is absent in the standard Langevin formulation.
\end{proposition}

Starting from the SGD update (Eq.~(\ref{E1})) and using the moment assumptions (Eq.~(\ref{E2})), the probability density evolves as
\begin{equation}
\label{eq:sgd-km-1d}
p_{n+1}(\theta)=\sum_{m=0}^\infty\frac{\eta^m}{m!}\nabla^m\Big(\mathbb{E}_L[(\nabla L(\theta))^m]\,p_n(\theta)\Big),
\end{equation}
which yields, to second order,
\begin{equation}
\label{eq:sgd-second-order-1d}
\begin{aligned}
p_{n+1}(\theta)
&= p_n(\theta)
 + \eta \nabla\Big(
   (\nabla \bar L(\theta))\,p_n(\theta)
   \Big)
\\
&\quad
 + \frac{1}{2}\eta^2 \nabla^2
 \Big(
   \big(D(\theta)+(\nabla \bar L(\theta))^2\big)
   p_n(\theta)
 \Big)
\\
&\quad
 + \mathcal{O}(\eta^3).
\end{aligned}
\end{equation}
By contrast, the Langevin approximation leads to
\begin{equation}
\label{eq:langevin-fp-1d}
\partial_t p(t,\theta)=\nabla\Big((\nabla \bar L(\theta))\,p(t,\theta)\Big)
+\frac{1}{2}\eta\nabla^2\Big(D(\theta)\,p(t,\theta)\Big).
\end{equation}

Comparing Eq.~\eqref{eq:sgd-second-order-1d} with Eq.~\eqref{eq:langevin-fp-1d}, we observe that the discrete expansion contains the additional second-order contribution
\[
\frac{1}{2}\eta^2 \nabla^2\Big((\nabla \bar L(\theta))^2 p_n(\theta)\Big),
\]
which is absent from the Langevin-based Fokker--Planck equation. Since this term is of the same formal order as the retained diffusion contribution, the Langevin approximation does not provide a consistent finite-step truncation of the discrete SGD dynamics.

A modified Langevin model,
\begin{equation}
    d\theta(t)=-\nabla\bar L(\theta(t))dt-\sqrt{\eta \big(D(\theta(t))+\big(\nabla \bar L(\theta)\big)^2\big)}dW_t,
    \label{E12}
\end{equation}
recovers the discrete result up to second order, but still differs at higher orders.

\paragraph{Toy example.}
To illustrate the practical effect of this discrepancy, consider
\begin{equation}
    L(\theta)=L_0+G\theta+\frac12 H\theta^2,
\end{equation}
where $G$ and $H$ are independent Gaussian random variables with moments
\begin{equation}
    \begin{aligned}
        \mathbb{E}_L[G] &= 0,
        &\qquad
        \mathbb{E}_L[G^2] &= d, \\
        \mathbb{E}_L[H] &= \lambda,
        &\qquad
        \mathbb{E}_L[H^2] &= \lambda^2+\Gamma .
    \end{aligned}
\end{equation}
Then
\begin{equation}
    \nabla \bar L(\theta)=\lambda\theta,\qquad D(\theta)=d+\Gamma\theta^2.
\end{equation}

For discrete SGD, the long-time variance is finite if
\[
\eta\Gamma<\lambda(2-\eta\lambda),
\]
and in that case
\begin{equation}
    \Pi_{n\to\infty}=\frac{\eta d}{\lambda(2-\eta\lambda)-\eta\Gamma}.
\end{equation}

For the standard Langevin approximation, the corresponding condition is only $\eta\Gamma<2\lambda$ with limiting variance
\begin{equation}
    \label{eq:langevin_variance}
    \Pi(t\to\infty)=\frac{\eta d}{2\lambda-\eta\Gamma}.
\end{equation}

This discrepancy is not caused by the toy model itself, but by the second-order terms omitted in the standard Langevin truncation. Indeed, if one instead uses the modified Langevin equation whose diffusion coefficient retains the additional $(\nabla \bar L)^2$ contribution from the discrete expansion, then the same calculation recovers the discrete SGD criterion
\[
\eta\Gamma<\lambda(2-\eta\lambda)
\]
and the limiting variance
\begin{equation}
    \Pi(t\to\infty)=
    \frac{\eta d}{\lambda(2-\eta\lambda)-\eta\Gamma}.
\end{equation}
Thus, the modified Langevin description \eqref{E12} agrees with discrete SGD at this order, whereas the standard Langevin approximation does not.

Therefore, in the regime $2\lambda-\eta\lambda^2<\eta\Gamma<2\lambda$ the standard Langevin approximation predicts a stationary state with finite variance, while the exact discrete SGD dynamics yields exponential growth of variance. This is a qualitative difference, not merely a quantitative correction.

Note that the modified Langevin equation \eqref{E12} predicts the same behavior as exact SGD.

\paragraph{Local analysis near a critical point.}
We now apply Eq.~\eqref{eq:FP} to SGD near a critical point of the average loss. Let $v$ be such a point. In a neighborhood of $v$, we approximate the minibatch loss by a quadratic form:
\begin{equation}
\label{eq:loss}
L(w)=L_0+\sum_{i=1}^d G_i (w_i - v_i)
+\frac{1}{2}\sum_{i,j=1}^d H_{ij}(w_i - v_i)(w_j - v_j)+\dots
\end{equation}
Here $G_i=\nabla_i L_n(v)$ are the stochastic gradient components and $H_{ij}$ is the stochastic Hessian at $v$.

Averaging over minibatches gives
\begin{equation*}
\mathbb{E}_L[L(w)]
=\mathbb{E}_L[L_0]
+\frac{1}{2}\sum_{i,j=1}^d \mathbb{E}_L[H_{ij}] (w_i - v_i)(w_j - v_j)+\dots,
\end{equation*}
since $\mathbb{E}_L[G_i]=0$. Thus, $v$ is a critical point of the average loss, although it need not be a minimum of each individual minibatch loss.

\paragraph{Mean and covariance of parameter trajectories.}
We characterize the parameter distribution by its mean, measured relative to $v$,
\begin{equation*}
\mu_i^n=\int (w_i - v_i)\, p_n(w)\, dw,
\end{equation*}
and covariance matrix
\begin{equation*}
\Pi_{ij}^n=\int (w_i-v_i)(w_j-v_j)\, p_n(w)\, dw - \mu_i^n \mu_j^n.
\end{equation*}
Substituting the quadratic approximation~\eqref{eq:loss} into Eq.~\eqref{eq:FP} yields a closed evolution equation for $\Pi_{ij}^n$ (Appendix~\ref{app:minimum}).

\paragraph{Mean-Hessian eigenbasis.}
The covariance dynamics simplify substantially in the eigenbasis of the mean Hessian,
\[
\mathbb{E}_L[H_{ij}] = \sum_{k=1}^d \lambda_k O_{ki}O_{kj},
\]
where $\lambda_k$ are the eigenvalues and $O$ is orthogonal. In this basis we define
\begin{equation}
\label{rotated}
\tilde H_{ij}=\sum_{k,l=1}^d O_{ik}O_{jl}H_{kl},\qquad
\tilde \Pi^n_{ij}=\sum_{k,l=1}^d O_{ik}O_{jl}\Pi^n_{kl}.
\end{equation}
Empirically, the rotated minibatch Hessian $\tilde H_{ij}$ is diagonally dominant, with fluctuations primarily affecting the diagonal entries due to SGD noise.

\paragraph{Assumptions.}
To obtain a tractable closed-form expression, we adopt the following assumptions (see Appendix~\ref{app:minimum}):
\begin{itemize}
\item independence of fluctuations of the components of $\tilde H_{ij}$;
\item approximate diagonality of $\mathbb{E}_L[\tilde H_{ij}^2]$;
\item small step size, $\eta \lambda^{\max} < 1$, where $\lambda^{\max}$ is the largest eigenvalue of the mean Hessian;
\item empirical relation between gradient covariance and Hessian:
\[
\mathbb{E}_L[G_i G_j]\approx \gamma\big(\mathbb{E}_L[H_{ij}]+\epsilon\big),
\]
with constant $\gamma$ and small $\epsilon$.
\end{itemize}
These assumptions reflect the empirical structure of minibatch noise and the standard finite-step stability condition.

The structure of the covariance dynamics can now be made explicit in the mean-Hessian eigenbasis.

\paragraph{Main result near a critical point.}
\begin{proposition}[Variance dynamics in the mean-Hessian eigenbasis]
\label{prop:variance}
Let $v$ be a critical point of the averaged loss, $\nabla \bar L(v)=0$, and assume the
local quadratic approximation
\[
L(w)=L_0+\sum_{i=1}^d G_i(w_i-v_i)
+\frac12\sum_{i,j=1}^d H_{ij}(w_i-v_i)(w_j-v_j).
\]
Let $\lambda_i$ be the eigenvalues of the mean Hessian $\mathbb E_L[H]$. Under the
assumptions stated above, the covariance matrix is approximately diagonal in the
mean-Hessian eigenbasis, and its diagonal entries satisfy
\begin{equation}
\label{key}
\tilde \Pi_{ii}^{n}
=
\eta\gamma
\frac{
1-\big(1-2\eta\lambda_i+\eta^2\mathbb E_L[\tilde H_{ii}^2]\big)^n
}{
2\lambda_i-\eta\mathbb E_L[\tilde H_{ii}^2]
}
(\lambda_i+\epsilon).
\end{equation}
Here $\gamma$ is independent of the learning rate, $\epsilon$ is a small regularizer,
and $n$ is the iteration index.
\end{proposition}

The derivation of Proposition~\ref{prop:variance} is given in Appendix~\ref{app:minimum}. We emphasize that this expression is approximate and relies on the assumptions stated above. In particular, the proportionality between gradient noise and curvature is an empirical observation rather than a general theoretical property.

\begin{corollary}[Diffusive and confined eigendirections]
\label{cor:regimes}
Under the assumptions of Proposition~\ref{prop:variance}, SGD separates into distinct
dynamical regimes in the mean-Hessian eigenbasis.

First, for nearly flat directions, $\lambda_i \approx 0$, the covariance grows
approximately linearly with time:
\[
\tilde\Pi_{ii}^n \propto n .
\]
More explicitly, before saturation and for $\eta n |\lambda_i| \ll 1$, the recursion
reduces to
\[
\tilde\Pi_{ii}^n \approx \gamma\eta^2 n\, (\lambda_i + \epsilon).
\]
These directions remain diffusive over the observed time horizon
and correspond to broad valleys of the loss landscape.

Second, for sufficiently sharp directions with positive curvature and
\[
\left|1-2\eta\lambda_i+\eta^2\mathbb E_L[\tilde H_{ii}^2]\right|<1,
\]
the covariance approaches a finite limiting value. In the small-step regime this limit is
\[
\tilde\Pi_{ii}\approx \frac12\eta\gamma.
\]

Finally, intermediate directions interpolate between these two limits and are described
by the full expression in Proposition~\ref{prop:variance}. Thus, within the local quadratic regime and over the observed time horizon, SGD near a critical point does not generally converge to a fully stationary distribution: sharp directions
are confined, while nearly-flat directions exhibit persistent spreading.
\end{corollary}

\section{Experiments}
\label{sec:experiments}
\paragraph{Small-scale validation.}
We first validate the assumptions and structural predictions of the theory in a small-scale NanoGPT/Shakespeare setting, where Hessian statistics and covariance dynamics can be measured directly. In particular, access to exact Hessians allows us to explicitly verify the assumptions underlying the theoretical derivation. 

Figure~\ref{fig:nano_te_appendix} compares the theoretical diagonal covariance profile from Eq.~\eqref{key} with the empirical covariance in the mean-Hessian eigenbasis. Since the overall multiplicative constant $\gamma$ is not fixed by the theory, the comparison is structural rather than absolute. We observe good agreement, including the predicted separation between saturating directions and directions that remain non-stationary over the observed time horizon. This supports both the assumptions and the resulting covariance dynamics derived in Proposition~\ref{prop:variance}.

\begin{figure}[t]
\centering
    \includegraphics[width = 0.48\textwidth]{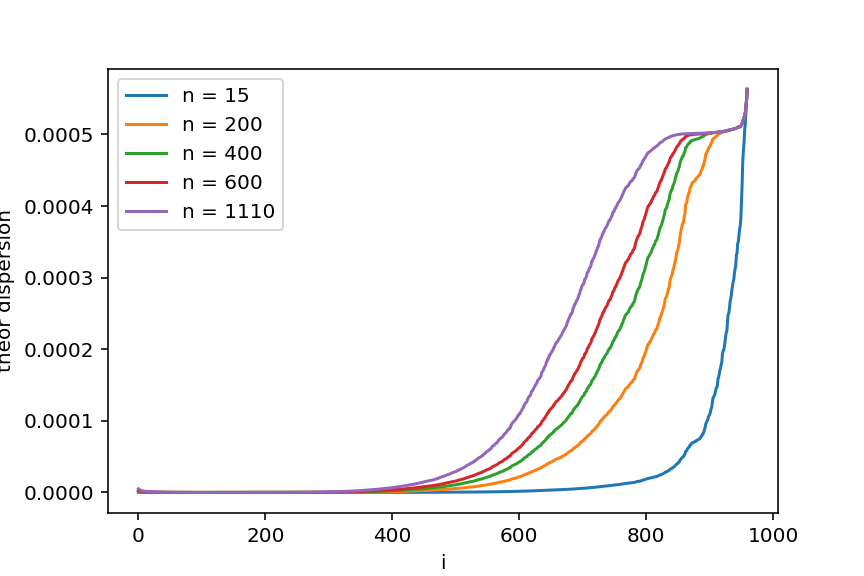}
    \includegraphics[width = 0.48\textwidth]{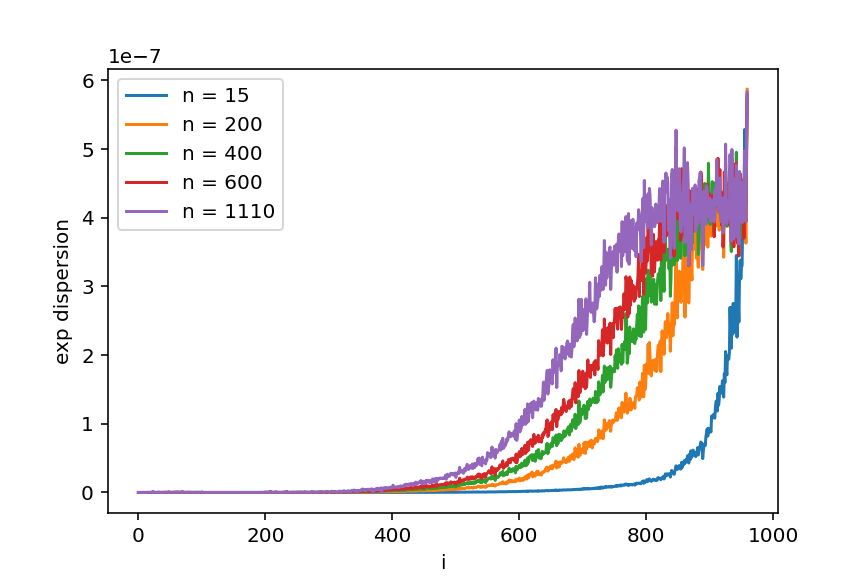}     
    \caption{
    \textbf{Small-scale NanoGPT/Shakespeare validation.}
    Left: theoretical prediction for the diagonal covariance profile from Eq.~\eqref{key}.
    Right: empirical measurement in the mean-Hessian eigenbasis.
    The comparison is structural, since the overall multiplicative constant $\gamma$ is not fixed by the theory.
    }
    \label{fig:nano_te_appendix}
\end{figure}

\paragraph{Empirical validation at scale.}
Having validated the assumptions in a controlled small-scale setting, we now test the variance prediction of Proposition~\ref{prop:variance} on a larger NanoGPT model with $6.6$M parameters trained on \\ WikiText-2.

\paragraph{Setup.}
We construct a reference point $w^*$ by SGD training followed by a short full-gradient refinement. We approximate the top-$20$ sharp eigendirections $v_i$ of the mean Hessian $\mathbb{E}_L[H]$ using stochastic Lanczos, and launch an ensemble of $N=50$ independent SGD trajectories from $w^*$ using sampling with replacement.

The empirical covariance is measured as
\begin{equation}
    \hat{\Pi}^n_{ii}
    =
    \mathrm{Var}_{j=1,\ldots,N}\!\left[
    \langle w_n^{(j)} - w^*,\, v_i\rangle
    \right],
\end{equation}
which provides a direct estimate of the theoretical quantity $\tilde{\Pi}_{ii}^n$ in Eq.~\eqref{key}. We evaluate three learning rates $\eta \in \{0.001, 0.005, 0.010\}$. A single $\hat{\gamma}$ is estimated at $\eta=0.001$ using the asymptotic relation $\tilde{\Pi}_{ii}^{\infty} \approx \tfrac12 \eta \gamma$ and then used without refitting for the other learning rates.

\begin{figure*}[t]
    \centering
    \includegraphics[width=0.85\textwidth]{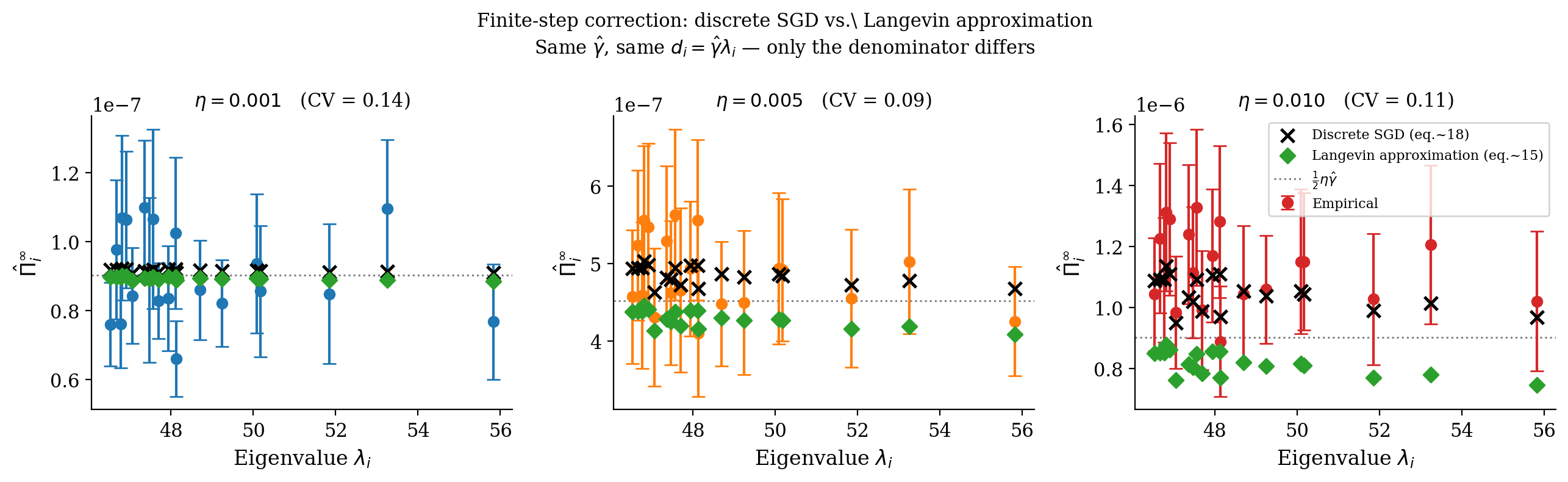}
    \caption{\textbf{Discrete SGD vs.\ Langevin approximation: saturation level (NanoGPT 6.6M).}
    Empirical plateau $\hat{\Pi}^\infty_i$ vs.\ Hessian eigenvalue $\lambda_i$ for the top-$20$ sharp directions and three learning rates.
    Circles: empirical means; crosses: discrete prediction (Eq.~\ref{key}); diamonds: Langevin prediction (Eq.~\eqref{eq:langevin_variance}).
    The Langevin approximation increasingly underestimates the plateau at larger $\eta$, while the discrete prediction remains consistent with the data, highlighting the importance of finite-step corrections.}
    \label{fig:big_level}
\end{figure*}

\paragraph{Results.}
Figure~\ref{fig:big_level} directly tests the stationary prediction of Proposition~\ref{prop:variance}. The empirical plateau $\hat{\Pi}_i^\infty$ is approximately independent of $\lambda_i$ and scales linearly with the learning rate $\eta$, in agreement with the theoretical prediction. Deviations from the Langevin approximation increase with $\eta$, consistent with the predicted finite-step corrections.

\paragraph{Discussion.}
Together, Figures~\ref{fig:nano_te_appendix} and~\ref{fig:big_level} support the theoretical picture: the assumptions hold in a controlled setting, and the resulting predictions remain accurate at scale. In particular, sharp directions are confined with a finite variance, while nearly flat directions remain diffusive over the observed time horizon. Additional implementation details are provided in Appendix~\ref{appendix:setups}.

\section{Limitations}

Our analysis relies on several simplifying assumptions and practical constraints. 
First, the theory is based on a local quadratic approximation near a reference point, which may not capture global dynamics or strongly non-smooth regions of the loss landscape. 
Second, we assume approximate independence and diagonality of minibatch Hessian fluctuations in the mean-Hessian eigenbasis; while supported empirically, this may not hold universally across architectures or training regimes. 
Third, the analysis is formulated for sampling with replacement, which differs from standard epoch-based training.

On the empirical side, exact validation is limited to small models where dense Hessians can be computed, while larger-scale experiments rely on low-dimensional diagnostics and primarily probe sharp directions. 
Finally, the comparison is partly structural: the framework does not determine certain constants (e.g., $\gamma$), and reproducibility is limited by the experimental pipeline and uncontrolled sources of randomness.

Extending the analysis beyond local regimes, relaxing structural assumptions, and improving large-scale validation remain important directions for future work.

\section{Conclusion}

In this work, we developed a discrete-time description of SGD dynamics by deriving a Fokker--Planck equation directly from the update rule. This approach avoids continuous-time approximations and makes explicit the role of minibatch-induced fluctuations at finite learning rate. In particular, it reveals correction terms that are not captured by standard Langevin-based models.

Using this framework, we analyzed SGD dynamics near critical points of the loss landscape. We showed that the evolution of parameter variance is governed by the local Hessian structure and exhibits distinct regimes across eigendirections. Within the local quadratic regime, directions with large curvature exhibit approximately stationary variance, while nearly-flat directions remain non-stationary and diffuse over time. This leads to a decomposition of SGD dynamics into confined and diffusive modes.

We provided empirical evidence for these predictions by comparing theoretical variance profiles with measurements from SGD trajectories in vision and language models. The observed behavior is consistent with the theoretical picture, with clear separation between saturating and non-saturating directions.

Overall, our results suggest that SGD can be more accurately understood as deterministic dynamics in a fluctuating loss landscape rather than as Brownian motion in a fixed potential, especially at finite learning rates. This perspective provides a unified explanation of several empirical phenomena and highlights the importance of discrete-time effects in understanding optimization dynamics.

\end{mainpart}

\begin{appendixpart}

\section{SGD Fokker-Planck equation and Langevin-like approach}
\label{app:Lang}
\paragraph{Proof of Proposition~\ref{prop:discrete-fp}.}
We prove the result by expanding the exact master equation (Eq.~(\ref{P})) in powers of $\eta$ and identifying the resulting terms up to second order.

In this appendix we derive the exact recursion for the parameter distribution induced by one SGD step and then obtain its second-order expansion in the learning rate. We then compare this discrete expansion with the standard Langevin-inspired surrogate dynamics and show that the two descriptions differ already at order $\eta^2$.

\subsection{Exact recursion from the master equation}

Our starting point is the master equation, which relates the parameter distribution at two successive iterations (Eq.~3):
\begin{equation}
p_{n+1}(w) = \int p_n(v) \,\mathbb{E}_L
\left[
\delta\left(w - v + \eta \nabla L(v)\right)
\right] dv.
\tag{20}
\end{equation}

We employ the representation of the delta function via its Fourier transform:
\begin{equation}
\delta(a) = \int \frac{dp}{(2\pi)^d} e^{ip^\top a}, \quad p \in \mathbb{R}^d.
\tag{21}
\end{equation}

This yields:
\begin{equation}
p_{n+1}(w) =
\int dv \, dp \, \frac{1}{(2\pi)^d}
p_n(v) e^{ip^\top(w-v)} \,
\mathbb{E}_L \left[e^{i\eta p^\top \nabla L(v)}\right].
\tag{22}
\end{equation}

We now expand the exponential in $\eta$:
\begin{equation}
p_{n+1}(w) =
\sum_{m=0}^{\infty} \frac{(i\eta)^m}{m!}
\sum_{i_1,\dots,i_m=1}^d
\int dv \, dp \, \frac{1}{(2\pi)^d}
p_n(v) e^{ip^\top(w-v)}
p_{i_1}\cdots p_{i_m}
\mathbb{E}_L\left[\nabla_{i_1}L(v)\cdots \nabla_{i_m}L(v)\right].
\tag{23}
\end{equation}

Using:
\begin{equation}
\int \frac{dp}{(2\pi)^d} e^{ip^\top(w-v)} p_{i_1}\cdots p_{i_m}
=
(-i)^m \nabla_{i_1}\cdots \nabla_{i_m} \delta(w-v),
\tag{24}
\end{equation}

we obtain the recursion:
\begin{equation}
p_{n+1}(w) =
\sum_{m=0}^{\infty} \frac{\eta^m}{m!}
\sum_{i_1,\dots,i_m=1}^d
\nabla_{i_1}\cdots \nabla_{i_m}
\left(
\mathbb{E}_L[\nabla_{i_1}L(w)\cdots \nabla_{i_m}L(w)] p_n(w)
\right).
\tag{25}
\end{equation}

\textbf{Important:} Equation (25) is an \emph{exact discrete identity} for SGD with independent minibatch sampling. No continuous-time approximation or truncation has been used.

\subsection{Second-order Fokker--Planck approximation}

To obtain a Fokker--Planck-type equation, we truncate the exact recursion (25) at second order in $\eta$.

This requires:
\begin{itemize}
\item sufficient smoothness of the loss,
\item existence of gradient moments,
\item small learning rate $\eta$.
\end{itemize}

The resulting second-order truncation is given in the main text.

\subsection{Langevin-inspired surrogate dynamics}

We now compare the above exact discrete expansion with the standard Langevin-inspired approximation of SGD.

Recall the SGD update:
\begin{equation}
w^i_{n+1} = w^i_n - \eta \nabla_i L(w_n).
\tag{26}
\end{equation}

The standard approach decomposes the loss into mean and fluctuations and models SGD as:
\begin{equation}
w^i_{n+1} =
w^i_n - \eta \nabla_i \bar L(w_n)
- \sqrt{\eta} \sum_{j=1}^d C^{ij}(w_n)\xi^j_n,
\tag{27}
\end{equation}
where $\xi_n$ is i.i.d. Gaussian noise:
\[
\mathbb{E}[\xi^i_n \xi^j_m] = \delta^{ij}\delta_{nm}.
\]

The covariance is matched via:
\begin{equation}
\sum_k C^{ik}(w)C^{kj}(w)
=
\eta \mathbb{E}_L[\nabla_i L(v)\nabla_j L(v)]
- \eta \nabla_i \bar L(w)\nabla_j \bar L(w)
=: W_{ij}(w).
\tag{28}
\end{equation}

The corresponding master equation becomes:
\begin{equation}
p_{n+1}(w) =
\int p_n(v)\,
\mathbb{E}_\xi
\left[
\delta\left(
w - v + \eta \nabla \bar L(v) + \sqrt{\eta}C(v)\xi_n
\right)
\right] dv.
\tag{29}
\end{equation}

Using Fourier representation:
\begin{equation}
p_{n+1}(w) =
\int dv \, dp \, \frac{1}{(2\pi)^d}
e^{ip^\top(w-v+\eta\nabla \bar L(v))}
\mathbb{E}_\xi
\left[e^{i\sqrt{\eta}p^\top C(v)\xi_n}\right]
p_n(v).
\tag{30}
\end{equation}

Averaging over Gaussian noise gives:
\begin{equation}
p_{n+1}(w) =
\int dv \, dp \, \frac{1}{(2\pi)^d}
e^{ip^\top(w-v)+i\eta p^\top \nabla \bar L(v) - \eta p^\top W(v)p}
p_n(v).
\tag{31}
\end{equation}

For small $\eta$ this yields:
\begin{equation}
p_{n+1}(w) =
p_n(w)
+ \eta \sum_k \nabla_k\left(p_n(w)\nabla_k \bar L(w)\right)
+ \frac{1}{2}\eta \sum_{k,l} \nabla_k\nabla_l\left(W_{kl}(w)p_n(w)\right)
+ \dots
\tag{32}
\end{equation}

\subsection[Finite-step mismatch at order eta squared]{Finite-step mismatch at order $\eta^2$}

We now compare Eq.~(32) with the exact expansion (25).

At first sight the expressions are similar. However, writing:
\[
W_{kl}(w)
=
\eta \mathbb{E}_L[\nabla_k L(w)\nabla_l L(w)]
- \eta \nabla_k \bar L(w)\nabla_l \bar L(w),
\]
we see that the diffusion term in Eq.~(32) is already of order $\eta^2$.

In contrast, the exact expansion (25) contains the second-order contribution:
\[
\frac{\eta^2}{2}
\nabla_k\nabla_l
\left(
\mathbb{E}_L[\nabla_k L(w)\nabla_l L(w)] p_n(w)
\right),
\]
which can be decomposed as:
\[
\mathbb{E}_L[\nabla_k L(w)\nabla_l L(w)]
=
D_{kl}(w) + \nabla_k \bar L(w)\nabla_l \bar L(w).
\]

Therefore, the exact discrete expansion contains the term:
\[
\frac{\eta^2}{2}
\nabla_k\nabla_l
\left(
\nabla_k \bar L(w)\nabla_l \bar L(w) p_n(w)
\right),
\]
which is \textbf{absent} from the Langevin-based recursion (32).

This discrepancy arises because the Langevin approximation retains a diffusion term of total order $\eta^2$ while discarding other contributions of the same order.

\subsection{Conclusion of the comparison}

Thus, the difference between discrete SGD and its Langevin approximation is not merely interpretational. The mismatch appears already at order $\eta^2$ and can lead to qualitatively different dynamics.

In contrast to the Langevin construction, the discrete formulation (25) keeps all contributions consistently organized in powers of $\eta$.

\section{Motion near the Landscape critical point: complete calculations }
\label{app:minimum}

In this appendix we derive the covariance dynamics used in the main text under a local quadratic approximation near a critical point of the averaged loss. 

The structure of the derivation is as follows. 
First, we introduce a quadratic approximation of the loss and derive closed equations for the mean and covariance of the parameter distribution. 
Next, we rotate these equations to the eigenbasis of the mean Hessian, which diagonalizes the deterministic part of the dynamics. 
Finally, we introduce simplifying assumptions that allow us to obtain explicit expressions for the variance and connect them to the main-text result.

Our starting point is the Fokker–Planck equation, which is
\begin{multline}
    p_{n+1}(w)=p_n(w)+\eta\sum_{k=1}^d \nabla_k\big(p_n(w)\,\nabla_k \bar L(w)\big)\\
    +\frac{1}{2}\eta^2\sum_{k,l=1}^d \nabla_k\nabla_l\Big(\mathbb{E}_L[\nabla_k L(w)\,\nabla_l L(w)]\,p_n(w)\Big)+\dots
\end{multline}
To study the local dynamics, we approximate the minibatch loss in a neighborhood of a critical point $w_c = v$ of the averaged loss, where $\nabla \bar L(v)=0$, by a quadratic expansion:
\begin{equation}
L(w)=L_0+\sum_{i=1}^d G_i (w_i-v_i)
+\frac{1}{2}\sum_{i,j=1}^d H_{ij}(w_i-v_i)(w_j-v_j)+\dots
\end{equation}
Here $L_0$, $G_i$, and $H_{ij}$ fluctuate across minibatches. The average loss is 
\begin{equation}
\mathbb{E}_L[L(w)]
=\mathbb{E}_L[L_0]
+\frac{1}{2}\sum_{i,j=1}^d \mathbb{E}_L[H_{ij}] (w_i-v_i)(w_j-v_j)+\dots
\end{equation}
so $\mathbb{E}_L[G_i]=0$. Thus, although individual minibatch losses may have nonzero gradients at $v$, the average loss has a stationary point at $v$. This distinction is important: stochasticity enters through fluctuations around this critical point. 
Under the quadratic approximation, the dynamics of the distribution can be fully characterized by its first two moments. Importantly, in this regime these moments form a closed system, which makes the analysis analytically tractable.

We define the mean displacement from the critical point as
\begin{equation}
\mu_i^n=\int (w_i-v_i)\, p_n(w)\, dw
\end{equation}
and the covariance matrix
\begin{equation}
\Pi_{ij}^n=\int (w_i-v_i)(w_j-v_j)\, p_n(w)\, dw - \mu_i^n \mu_j^n.
\end{equation}

Substituting the quadratic approximation into the Fokker–Planck equation and integrating by parts, one obtains a closed system of recursion relations for the mean and covariance:
\begin{equation}
    \mu_{n+1}^i=\mu_n^i-\eta\sum_{j=1}^d \mathbb{E}_L[H_{ij}]\,\mu_n^j,
    \label{mu_eq}
\end{equation}
\begin{equation}
    \Pi_{n+1}^{ij}
    =\Pi_n^{ij}
    -\eta\sum_{k=1}^d\Big(\Pi_n^{ik}\,\mathbb{E}_L[H_{kj}]
    +\mathbb{E}_L[H_{ik}]\,\Pi_n^{kj}\Big)
    +\eta^2\sum_{k,l=1}^d \mathbb{E}_L[H_{ik}H_{jl}]\,\Pi_n^{kl}
    +\eta^2\Lambda_n^{ij},
    \label{Pi_eq}
\end{equation}
where
\begin{equation}
    \Lambda_n^{ij}
    =
    \mathbb{E}_L\!\left[
    \Big(G_i+\sum_{k=1}^d H_{ik}\mu_n^k\Big)
    \Big(G_j+\sum_{l=1}^d H_{jl}\mu_n^l\Big)
    \right]
    -\sum_{k,l=1}^d
    \mathbb{E}_L[H_{ik}]\,\mathbb{E}_L[H_{jl}]\,\mu_n^k\mu_n^l.
\end{equation}
The covariance dynamics derived above contain many coupled terms between different coordinates. To simplify the analysis, we now pass to the eigenbasis of the mean Hessian.  Specifically, we diagonalize the matrix $\mathbb{E}_L[H_{ij}]$ as \[ \mathbb{E}_L[H_{ij}] = \sum_{k=1}^d \lambda_k O_{ki}O_{kj}, \] where $O_{ij}$ is an orthogonal matrix and $\lambda_i$ are the eigenvalues of $\langle H_{ij}\rangle_{L}$.  This transformation diagonalizes the deterministic part of the dynamics and provides a natural coordinate system in which different curvature directions can be analyzed independently. In rotated basis
\begin{equation}
    \tilde G_i=\sum_{k=1}^d O_{ik}G_k,\qquad
    \tilde H_{ij}=\sum_{k,l=1}^d O_{ik}O_{jl}H_{kl},\qquad
    \tilde\mu_n^i=\sum_{k=1}^d O_{ik}\mu_n^k,\qquad
    \tilde\Pi_n^{ij}=\sum_{k,l=1}^d O_{ik}O_{jl}\Pi_n^{kl}.
\end{equation}
and the equations ~\eqref{mu_eq},~\eqref{Pi_eq} have the form
\begin{equation}
    \tilde\mu_{n+1}^i=\tilde\mu_n^i-\eta\lambda_i\tilde\mu_n^i.
\end{equation}
\begin{equation}
    \tilde\Pi_{n+1}^{ij}
    =\tilde\Pi_n^{ij}
    -\eta(\lambda_i+\lambda_j)\tilde\Pi_n^{ij}
    +\eta^2\sum_{k,l=1}^d \mathbb{E}_L[\tilde H_{ik}\tilde H_{jl}]\,\tilde\Pi_n^{kl}
    +\eta^2\tilde\Lambda_n^{ij}.
\end{equation}
To make further analytical progress, we introduce the following simplifying assumptions on the statistics of minibatch fluctuations in the rotated basis.  First, we assume that different components of the gradient fluctuations are uncorrelated:
\begin{equation}
    \mathbb{E}_L[\tilde G_i\tilde G_j]=d_i\delta_{ij},\qquad
    \mathbb{E}_L[\tilde H_{ij}]=\lambda_i\delta_{ij},
\end{equation}

Second, we assume that the Hessian fluctuations are weakly correlated and can be expressed through their second moments:
\begin{equation}
    \mathbb{E}_L[\tilde H_{ik}\tilde H_{jl}]
    =\lambda_i\lambda_j\delta_{ik}\delta_{jl}
+\Gamma_{ik}\big(\delta_{ij}\delta_{kl}+\delta_{il}\delta_{jk}-\delta_{ij}\delta_{ik}\delta_{jl}\big).
\end{equation}
Here $\Gamma_{ij}$ characterizes the variance of fluctuations of the $(i,j)$-th component of the Hessian in the rotated basis. We first solve the recursion for the mean. Since the equation is linear and diagonal in the rotated basis, it can be solved explicitly, yielding \[ \tilde \mu^i_{n}=(1-\eta\lambda_i)^n\tilde\mu^i_{0}, \] where $\tilde\mu^i_{0}$ is the initial displacement.
Substituting this solution for the mean into the definition of $\tilde\Lambda_n^{ij}$, we obtain
\begin{equation}
    \tilde\Lambda_n^{ij}
    =(1-\eta\lambda_i)^n(1-\eta\lambda_j)^n(1-\delta_{ij})\Gamma_{ij}\tilde\mu_0^i\tilde\mu_0^j
    +\delta_{ij}\sum_{k=1}^d \Gamma_{ik}(1-\eta\lambda_k)^{2n}(\tilde\mu_0^k)^2
    +d_i\delta_{ij}.
\end{equation}
We now turn to the covariance dynamics. We begin with the off-diagonal elements ($i\neq j$), for which the recursion simplifies:
\begin{equation}
    \tilde\Pi_{n+1}^{ij}
    =\Big((1-\eta\lambda_i)(1-\eta\lambda_j)+\eta^2\Gamma_{ij}\Big)\tilde\Pi_n^{ij}
    +\eta^2(1-\eta\lambda_i)^n(1-\eta\lambda_j)^n\Gamma_{ij}\tilde\mu_0^i\tilde\mu_0^j.
\end{equation}
This linear recursion can be solved explicitly, yielding
\begin{equation}
    \tilde\Pi_n^{ij}
    =\Big(\big((1-\eta\lambda_i)(1-\eta\lambda_j)+\eta^2\Gamma_{ij}\big)^n
    -(1-\eta\lambda_i)^n(1-\eta\lambda_j)^n\Big)\tilde\mu_0^i\tilde\mu_0^j.
    \label{sol-nondiag}
\end{equation}
For the diagonal elements one has the equation
\begin{equation}
    \tilde\Pi_{n+1}^{ii}
    =(1-\eta\lambda_i)^2\tilde\Pi_n^{ii}
    +\eta^2\sum_{k=1}^d \Gamma_{ik}\Big(\tilde\Pi_n^{kk}+(1-\eta\lambda_k)^{2n}(\tilde\mu_0^k)^2\Big)
    +\eta^2 d_i.
\end{equation}
This equation is more complicated, however it can be solved in the matrix form. Let us introduce the diagonal matrix $\hat L$ with elements $L_{ij}=(1-\eta\lambda_i)^2\delta_{ij}$ and the matrix $\hat \Gamma$ with elements $\Gamma_{ij}$. Then the solution is
\begin{equation}
    \label{sol-diag}
    \begin{aligned}
        \tilde\Pi_n^{ii}
        &=\sum_{k=1}^d \big[(\hat L+\eta^2\hat\Gamma)^n-\hat L^n\big]_{ik}(\tilde\mu_0^k)^2 \\
        &\quad + \eta^2\sum_{k=1}^d
        \left[\frac{1-(\hat L+\eta^2\hat\Gamma)^n}{1-\hat L-\eta^2\hat\Gamma}\right]_{ik} d_k.
    \end{aligned}
\end{equation}
Solutions \eqref{sol-nondiag},\eqref{sol-diag} have two contributions to the variance matrix $\tilde\Pi^{ij}_{n}$. The first one is related to the initial conditions, whereas the second one is diagonal (in the basis chosen). Now let us analyze this general result in context of numerical experiments made by us.

Typically in our experiments the fluctuation matrix $\eta^2\Gamma$ are rather small (but not negligible) and the matrix $\Gamma_{ij}$ itself are nearly diagonal. We observe that one can safely neglect the non-diagonal elements of $\Gamma_{ij}$. Also one can check, that the term containing initial condition are unimportant. Then the variance matrix $\tilde \Pi_{ij}^{n}$ is diagonal in the basis composed of $\langle H_{ij}\rangle_{L}$ eigenvectors. This is supported by our experiments (see the main text). In other words in this basis all the directions evolve independently and their behavior depends on the  eigenvalues $\lambda_i$. For near-zero eigenvalues the $(1-\eta\lambda_i)^n$ quantity might be still near $1$ up to step $n$, so one can expand the variance in $\eta\lambda_i$. After the simple algebra we obtain $\Pi^{ii}_{n}\approx \eta^2 n d_k$ if $\eta n|\lambda_i|\ll 1$. These directions are non-stationary and the dynamics in these directions remain diffusive. Although the effective diffusion coefficient $\eta d_k$ turns out to be very small, the large number of these nearly-flat directions leads to a noticeable divergence of individual trajectories. In the directions with large eigenvalues $\eta n\lambda_i\gg 1$ the system reaches  the stationary state given by
\begin{equation}
    \Pi_n^{ii}\approx \frac{\eta d_i}{2\lambda_i-\eta\lambda_i^2-\eta\Gamma_{ii}}.
\end{equation}
The behavior in the other (intermediate) directions is in between these two limiting cases. 

We also observe empirically that, for directions with positive eigenvalues, $d_i$ is approximately proportional to $\lambda_i$. In other words $\mathbb{E}_L[G_i G_j]\approx \gamma\big(\mathbb{E}_L[H_{ij}]+\epsilon_{ij}\big)$, where the matrix $\epsilon_{ij}$ is small and necessary only for the regularization of negative eigenvalues. $\gamma$ is a some constant, which is independent of the learning rate. We propose a possible mechanism of this phenomena below, but now we substitute this relation into the result for large $\lambda_i$ directions
\begin{equation}
    \Pi_{ii}^n\approx \frac{\eta\gamma\lambda_i}{2\lambda_i-\eta\lambda_i^2-\eta\Gamma_{ii}}
    \approx \frac{\eta\gamma}{2}.
\end{equation}
We obtain the following picture. Near the critical point, there are many directions with small eigenvalues in which the system moves chaotically, resulting in divergence of trajectories. Whereas in the the rigid direction there is something like the evolution to the stationary state, but the width of this stationary distribution is almost the same for all (rigid) directions, i.e. independent of $\lambda_i$. For the intermediate directions the variance depends on the iteration number $n$. Under the assumptions stated above it is necessary to consider only diagonal elements of matrix $\hat L+\eta^2\hat\Gamma$, which are 
\begin{equation}
    (1-\eta\lambda_i)^2+\eta^2\Gamma_{ii}
    =1-2\eta\lambda_i+\eta^2\mathbb{E}_L[\tilde H_{ii}^2].
\end{equation}
The last equality follows from the definition of $\Gamma_{ij}$. Putting everything together one can reproduce the main result of our paper from the equation \eqref{sol-diag}
\begin{equation}
    \tilde\Pi_{ii}^{n}
    =\eta\gamma
    \frac{1-\big(1-2\eta\lambda_i+\eta^2\mathbb{E}_L[\tilde H_{ii}^2]\big)^n}
    {2\lambda_i-\eta\mathbb{E}_L[\tilde H_{ii}^2]}
    (\lambda_i+\epsilon).
\end{equation}
This result naturally explains the inverse Einstein relation discussed earlier in the literature. It is just the non-stationary effect. The effective fluctuations in the direction of small $\lambda_i$ are small and the system does not evolve in this directions resulting in the small distribution width at a finite number of steps.

At the end of this appendix let us make a few comments about the origin of $\langle G_i G_j\rangle_{L}\sim \langle H_{ij}\rangle_{L}$ relation. One of the possible mechanisms is the following. Consider the minibatch loss near the critical point in the form
\begin{equation}
    L(w)=\frac{1}{N_B}\sum_{k=1}^{N_B}\Phi\!\left(\sum_{i=1}^d b_i^{(k)}(w_i-v_i)\right).
\end{equation}
where $N_B$ is a minibatch size and $\vec b^{(k)}$ is a random vector which is attributed to the given sample. $\Phi(x)$ - is some function. Expanding it around the critical point one obtains
\begin{multline}
    L(w)\approx \Phi(0)
    +\frac{1}{N_B}\Phi'(0)\sum_{k=1}^{N_B}\sum_{i=1}^d b_i^{(k)}(w_i-v_i) \\
    +\frac{1}{2N_B}\Phi''(0)\sum_{k=1}^{N_B}\sum_{i,j=1}^d
    b_i^{(k)}b_j^{(k)}(w_i-v_i)(w_j-v_j)+\dots
\end{multline}
From this one can identify
\begin{equation}
    G_i=\frac{1}{N_B}\Phi'(0)\sum_{k=1}^{N_B} b_i^{(k)},\qquad
    H_{ij}=\frac{1}{N_B}\Phi''(0)\sum_{k=1}^{N_B} b_i^{(k)}b_j^{(k)}.
\end{equation}
so the relation
\begin{equation}
    \mathbb{E}_b[G_i G_j]
    =\frac{1}{N_B}\big(\Phi'(0)\big)^2\mathbb{E}_b[b_i b_j]
    =\frac{\big(\Phi'(0)\big)^2}{N_B\Phi''(0)}\,\mathbb{E}_b[H_{ij}].
\end{equation}
holds. This is only an illustrative example, but it captures the basic mechanism. The fluctuations of both the gradient and the Hessian originate from the same source, so they may share the same information about the randomness of the system and be proportional to the same matrix.

\section{Experimental setup}
\label{appendix:setups}

The code repository with implementation details and scripts for reproducing the experiments is available at:
\url{https://github.com/brain-lab-research/SGDiffusion}.

This section provides implementation details for the experiments used in the paper. 
The small-scale MNIST and Shakespeare experiments allow exact dense Hessian computation and are used to validate the structural assumptions behind Proposition~\ref{prop:variance}: approximate diagonalization in the mean-Hessian eigenbasis, separation into sharp and flat regimes, and agreement between the predicted and measured covariance profiles.

The large-scale NanoGPT experiment in Section~\ref{sec:big_model} complements these small-scale diagnostics by testing the sharp-direction saturation law on a $6.6$M-parameter model, where only Hessian-vector products are feasible.

Across the small-scale experiments, we use the same protocol:
\begin{enumerate}
    \item train a model with SGD until the loss is substantially reduced;
    \item refine the resulting checkpoint with full-gradient descent to obtain a reference point $w^*$;
    \item compute Hessian information near $w^*$ and rotate SGD trajectories into the mean-Hessian eigenbasis;
    \item compare the measured covariance with the theoretical prediction from Eq.~\eqref{key}.
\end{enumerate}

\subsection{Small-scale MNIST experiment}

We first consider MNIST~\citep{lecun2010mnist} with a compact MLP containing $386$ parameters. This model is small enough to compute dense minibatch Hessians exactly. After SGD training and GD refinement, we use the resulting point $w^*$ as the reference point for local analysis. We then compute minibatch Hessians near $w^*$ and form the mean Hessian $\mathbb{E}_L[H]$. Its eigenvalue spectrum is shown in Fig.~\ref{fig:mlp_lambda} (left).

\begin{figure}[t]
\centering
	\includegraphics[width=0.55\textwidth]{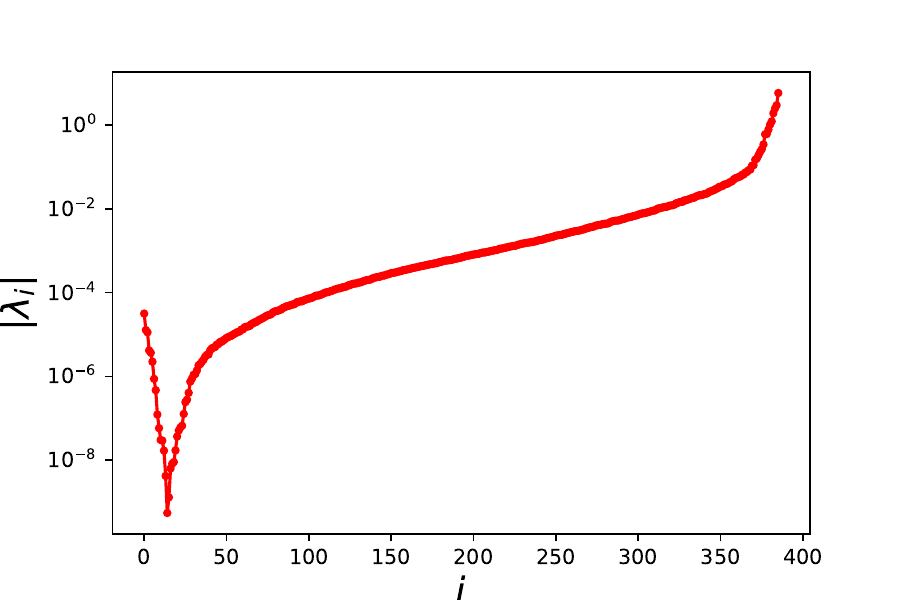}
    \hfill
    \includegraphics[width=0.4\textwidth]{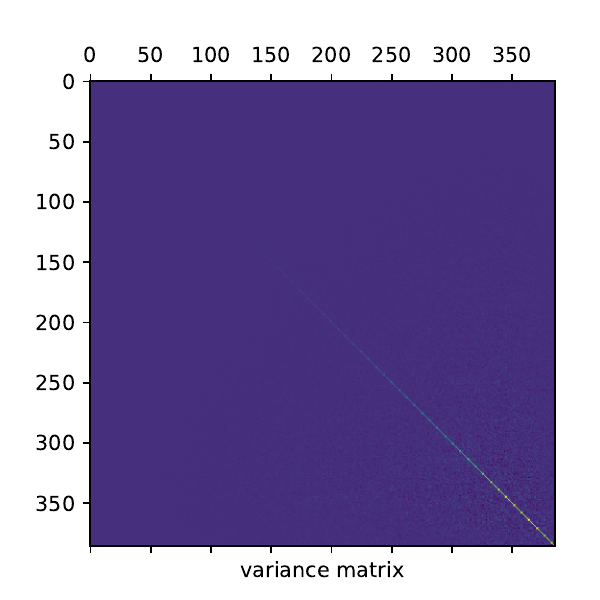}	
	\caption{MLP experiment. Left: eigenvalue spectrum of the mean Hessian. 
	Right: variance matrix of SGD trajectories in the mean-Hessian eigenbasis.}
	\label{fig:mlp_lambda}
\end{figure}

The spectrum reveals that most eigenvalues are clustered near zero or slightly negative, with only a few directions exhibiting significant positive curvature ($\lambda_{\max}\approx 6$). In this local region, the spectrum is dominated by wide, nearly flat directions, together with a small number of sharper directions.

To connect the spectrum with SGD dynamics, we launch $1000$ independent SGD trajectories from $w^*$ and project the displacements $w_n-w^*$ onto the eigenvectors of $\mathbb{E}_L[H]$. The resulting covariance matrix in this basis is shown in Fig.~\ref{fig:mlp_lambda} (right). Its near-diagonal structure supports the assumption that the mean-Hessian eigenbasis approximately decouples the dominant fluctuation modes.

\begin{figure}[t]
\centering
	\includegraphics[width = 0.48\textwidth]{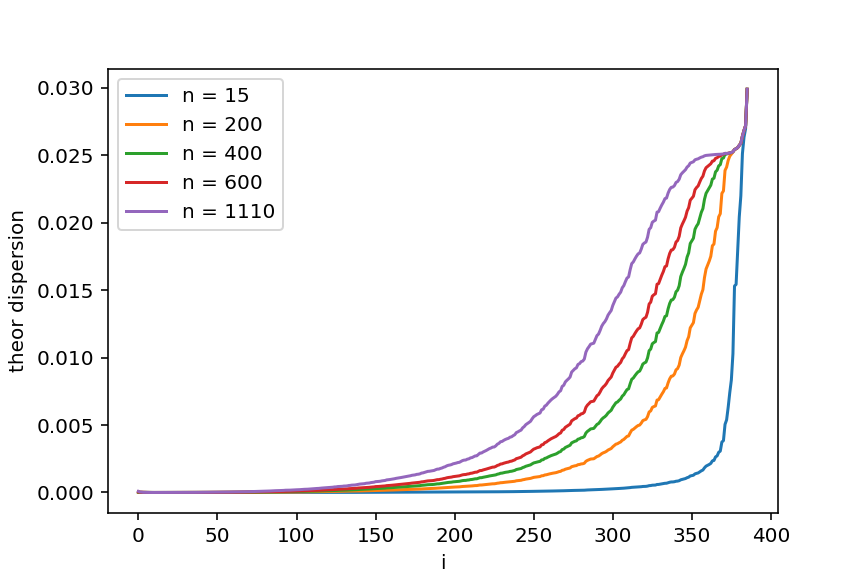}
    \includegraphics[width = 0.48\textwidth]{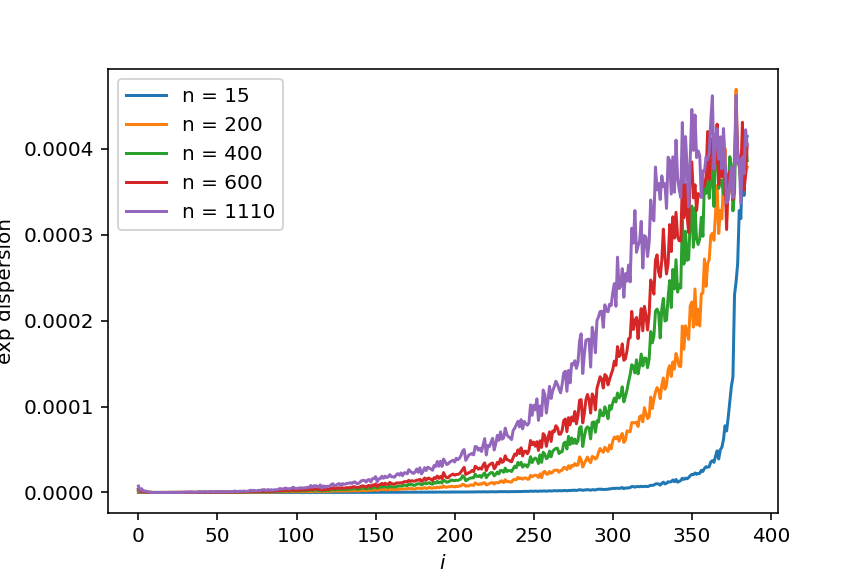}		
	\caption{MLP experiment. Left: theoretical prediction for the diagonal elements of the variance matrix from Eq.~\eqref{key}.
Right: experimental measurement of the same quantity. Note that the theory does not predict the value of the overall multiplicative constant $\gamma$; therefore, the magnitudes of the theoretical and experimental curves do not coincide.}
	\label{fig:mlp_te}
\end{figure}

Figure~\ref{fig:mlp_te} compares the predicted diagonal covariance profile from Eq.~\eqref{key} with the measured profile in the same mean-Hessian eigenbasis. Since the overall factor $\gamma$ is not predicted by the theory, the comparison is primarily structural. The predicted profile captures the qualitative separation between directions that saturate and directions that remain non-stationary over the observed horizon.

\subsection{Small-scale Shakespeare experiment}

To test whether the same phenomena appear beyond vision tasks, we repeat the analysis on a text modeling task. We train a simplified GPT ~\citep{Radford2018ImprovingLU} model on the Shakespeare corpus ~\citep{karpathy2015charrnn}.  
Here we use a simplified symbolic prediction task in which each character is shifted by one position in the alphabet, similar to a Caesar cipher~\citep{caesarCipherWiki}. This keeps the problem simple enough for full Hessian analysis while preserving the sequential structure of the data.
\begin{figure}[t]
	\includegraphics[width=0.55\textwidth]{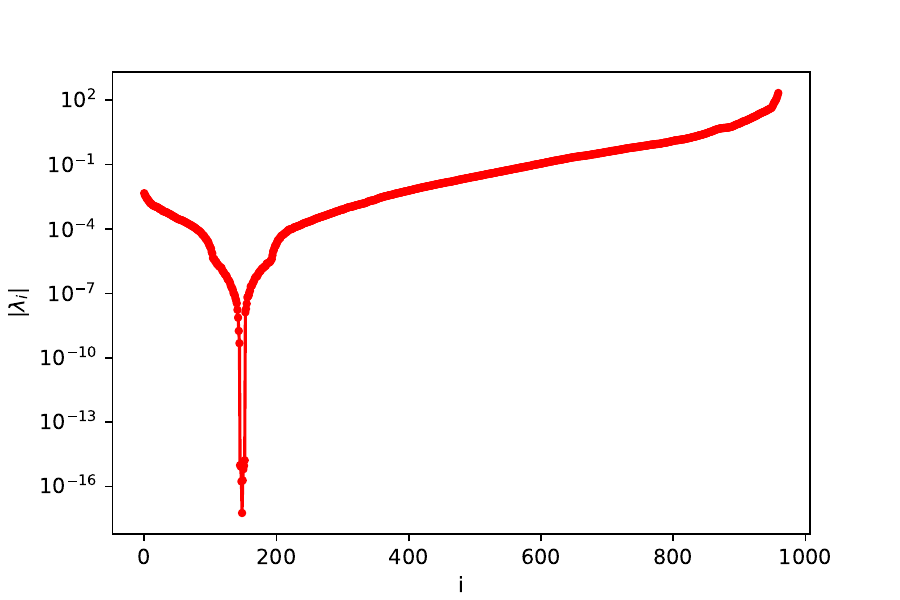}
    \hfill
    \includegraphics[width=0.4\textwidth]{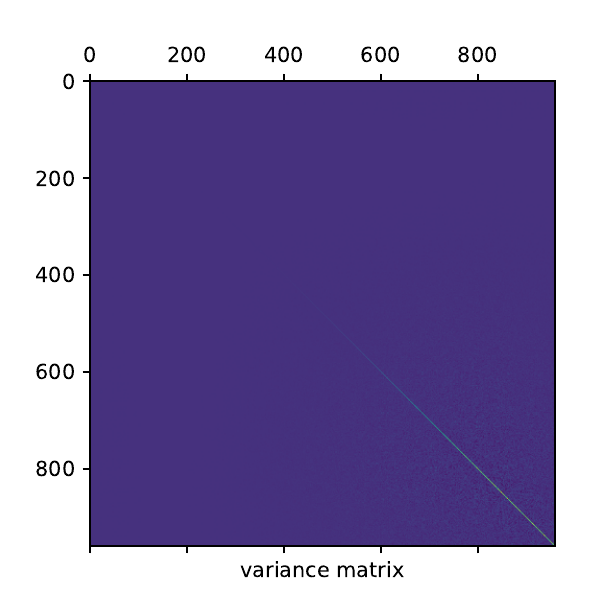}	
	\caption{NanoGPT language model. Left: eigenvalue spectrum of the mean Hessian.  
	Right: variance matrix of parameter trajectories in the mean-Hessian eigenbasis.}
	\label{fig:nano_lambda}
\end{figure}
For the language model, we repeated the same set of experiments as conducted for the MLP, with learning rate $\eta = 0.001$. The spectral structure differs in scale — the maximum eigenvalue is substantially larger, $\lambda_{\max}\approx 217$ (Fig.~\ref{fig:nano_lambda}, left). Nevertheless, as in the vision case, most eigenvalues are near zero, indicating that the loss landscape is again dominated by wide flat valleys.  

The variance matrix (Fig.~\ref{fig:nano_lambda}, right) is again close to diagonal in the Hessian eigenbasis. This shows that parameter fluctuations decouple along eigendirections, and the distinction between diffusive and rigid modes derived in the previous section applies equally well here. Example trajectories confirming these behaviors are provided in Section~\ref{sec:trajectories}.  
Taken together, these results show that the structural assumptions of the theory—approximate diagonalization in the mean-Hessian eigenbasis and separation into sharp and flat regimes—also hold in a language-model setting.

\subsection{SGD Trajectories in Different Eigendirections}
\label{sec:trajectories}

To illustrate the distinct behaviors predicted by our theoretical framework, we examine individual SGD trajectories projected onto representative eigendirections of the mean Hessian. We report results for both the MLP (vision) experiment (Fig.~\ref{fig:mlp_dir}) and the Shakespeare (language) model (Fig.~\ref{fig:nano_dir}).

\begin{figure}[t]
    \centering
	\includegraphics[width = 0.85\textwidth]{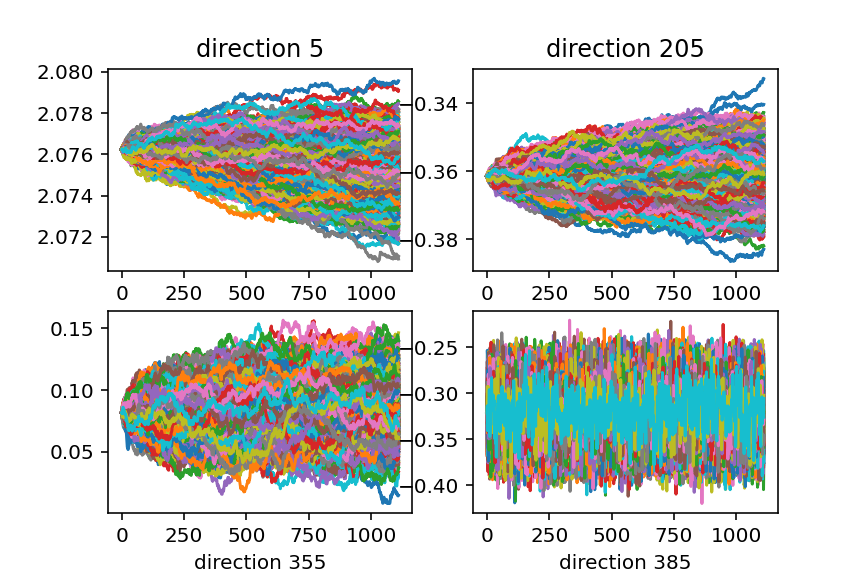}		
	\caption{MLP experiment: trajectories in distinct eigendirections of the mean Hessian. 
	\textbf{Top:} flat directions ($\lambda \approx 0$) show diffusive motion, with variance increasing over time.  
	\textbf{Bottom:} curved directions ($\lambda > 0$) remain confined, with variance saturating at a finite value.}
	\label{fig:mlp_dir}
\end{figure}

\begin{figure}[t]
    \centering
	\includegraphics[width = 0.85\textwidth]{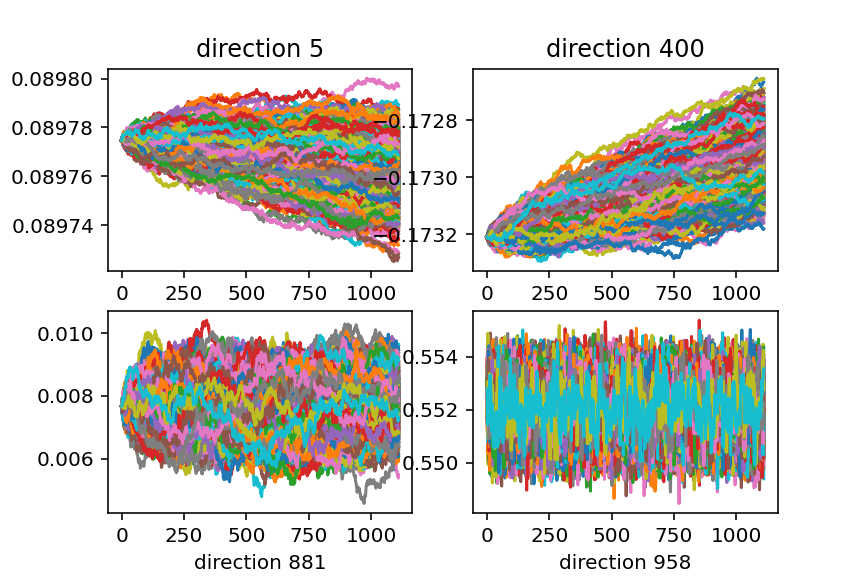}		
	\caption{Shakespeare model: trajectories in distinct eigendirections of the mean Hessian. 
	\textbf{Top:} flat directions exhibit diffusive behavior.  
	\textbf{Bottom:} curved directions show confined dynamics, consistent with the MLP case.}
	\label{fig:nano_dir}
\end{figure}

\paragraph{Diffusive directions (top panels).}  
For both vision and language tasks, in eigendirections corresponding to near‑zero or slightly negative eigenvalues, SGD trajectories display unconfined behavior. Variance grows steadily in time, consistent with diffusion along valley floors and in agreement with the Fokker–Planck analysis.  

\paragraph{Rigid directions (bottom panels).}  
In directions with large positive curvature, trajectories remain localized around the minimum. The variance quickly saturates to a finite value controlled by the balance between curvature and minibatch noise. In sharp positive-curvature directions, the variance saturates to a finite value. Under the empirical relation $d_i \propto \lambda_i$, this plateau becomes approximately independent of $\lambda_i$, consistent with the inverse variance--flatness relation discussed in the main text.

\paragraph{Cross‑task consistency.}  
The similarity of behaviors between MLP and Shakespeare supports the view that this decomposition is not specific to a single architecture or data domain: the decomposition of SGD dynamics into diffusive and rigid modes is not tied to the specific architecture or data domain, but instead appears to be a general property of optimization in high‑dimensional loss landscapes.

\subsection{Large-Scale Validation: NanoGPT 6.6M on WikiText-2}
\label{sec:big_model}

This section provides implementation details for the large-scale validation experiment presented in Section~\ref{sec:experiments}. In particular, we describe the training protocol, spectral estimation, and trajectory measurements underlying Fig.~\ref{fig:big_level}.

\paragraph{Model and data.}
We use a NanoGPT transformer with embedding dimension $64$, $2$ attention heads, $4$ layers, context length $256$, and MLP ratio $4$. The model is trained on WikiText-2 tokenized with the GPT-2 BPE vocabulary. We use the standard train/validation split.

\paragraph{Training protocol.}
Starting from a random initialization, we train with SGD using learning rate $\eta_{\mathrm{train}} = 0.05$, minibatch size $B = 32$, and sampling with replacement. Training proceeds until the validation loss plateaus: we stop when the mean loss over the most recent $200$ steps improves by less than $10^{-3}$ relative to the preceding window. This occurs after approximately $5{,}150$ SGD steps at validation loss $6.63$. We save the resulting checkpoint $w_{\mathrm{SGD}}$.

Starting from $w_{\mathrm{SGD}}$, we apply $100$ steps of full-gradient descent with learning rate $0.01$, accumulating gradients over the entire training set at each step. The result is our reference point $w^*$.

After $100$ full-gradient refinement steps, the full-batch gradient norm 
at $w^*$ satisfies $\|\nabla \bar{L}(w^*)\|_2 = 6.0 \times 10^{-2}$, with 
relative norm $\|\nabla \bar{L}(w^*)\|_2 / \|w^*\|_2 = 1.0 \times 10^{-3}$. 
As a measure of proximity to a critical point, we report 
$\|\nabla \bar{L}(w^*)\|_2 / \lambda_{\max} = 1.1 \times 10^{-3} \ll 1$, 
confirming that the gradient is small relative to the curvature scale and 
that the local quadratic approximation underlying Proposition\ref{prop:variance} is 
appropriate.

\paragraph{Spectral analysis via stochastic Lanczos.}
Since exact dense Hessian computation is infeasible at this scale, we apply 
stochastic Lanczos with full reorthogonalization (every $10$ iterations) to 
approximate the top-$K = 20$ eigenvectors of $\mathbb{E}[H]$ at $w^*$. Each 
Hessian-vector product is averaged over $5$ independent minibatches of size 
$4$, and we run $200$ Lanczos iterations in total. This yields eigenvalues 
$\lambda_i \in [46.5,\, 55.8]$, all strictly positive. We note that $w^*$ is 
not a true minimum of the loss — as is typical for large neural networks, the 
Hessian at a post-training checkpoint has negative eigenvalues, indicating 
residual saddle structure. This does not affect the validity of the analysis 
in sharp directions, where Proposition\ref{prop:variance} applies, but prevents us from 
directly probing the diffusive regime at this scale.

\paragraph{Local statistics at $w^*$.}
We estimate the gradient noise covariance $K_{ij} = \mathbb{E}[\tilde{G}_i 
\tilde{G}_j]$ and the diagonal second moment $\mathbb{E}[\tilde{H}_{ii}^2]$ 
projected onto the top-$20$ sharp eigenvectors. $K_{ij}$ is computed by 
averaging outer products of per-minibatch gradients over $N_{\mathrm{grad}} = 
500$ minibatches of size $B = 32$. $\mathbb{E}[\tilde{H}_{ii}^2]$ is estimated 
by averaging squared Hessian-vector products $(v_i^\top H_n v_i)^2$ over 
$N_{H_2} = 300$ independent minibatches of size $4$; the reduced minibatch 
size is due to memory constraints, as the double-backward pass required for 
HVP computation increases peak GPU memory by approximately $2$--$3\times$ 
relative to a standard forward pass.

\paragraph{Estimation of $\gamma$.}
We estimate $\gamma$ from the saturation level of the parameter variance ensemble at $\eta = 0.001$, where finite-step corrections $\eta\,\mathbb{E}[\tilde{H}_{ii}^2]$ are negligible relative to $2\lambda_i$ (below $0.1\%$), so that $\hat{\Pi}^\infty_{ii} \approx \frac{1}{2}\eta\gamma$ holds to high accuracy. Specifically,
\begin{equation}
    \hat{\gamma} = \frac{2}{\eta}\,\overline{\hat{\Pi}^\infty},
\end{equation}
where $\overline{\hat{\Pi}^\infty}$ is the mean empirical plateau averaged over the top-$20$ sharp directions and the last $200$ trajectory steps. This yields $\hat{\gamma} = 1.81 \times 10^{-4}$. As an independent check, a weighted least-squares regression of $d_i \approx \gamma \lambda_i$ over sharp directions gives $\hat{\gamma}_{\mathrm{WLS}} = 1.47 \times 10^{-4}$ with coefficient of variation $\mathrm{CV} = 0.23$, consistent with the saturation estimate within the expected uncertainty.

\paragraph{SGD trajectory ensemble.}
We generate $N = 50$ independent SGD trajectories from $w^*$, each using sampling with replacement and minibatch size $32$. At each step $n$, we project the parameter displacement $w_n - w^*$ onto all $20$ sharp eigenvectors and record the scalar projections. We repeat this for three learning rates $\eta \in \{0.001,\, 0.005,\, 0.010\}$, running $T = 500$ steps for $\eta \in \{0.001, 0.010\}$ and $T = 1000$ steps for $\eta = 0.005$. The empirical variance in direction $i$ at step $n$ is estimated as
\begin{equation}
    \hat{\Pi}^n_{ii} = \mathrm{Var}_{j=1,\ldots,N}\!\left[\langle w_n^{(j)} - w^*,\, v_i\rangle\right],
\end{equation}
where the variance is taken across the $N$ trajectory realizations, providing a direct test of the predicted variance dynamics (see Fig.~\ref{fig:big_dynamics}).

\paragraph{Results.}
The main empirical result is shown in Fig.~\ref{fig:big_level} in the main text. It demonstrates that the saturation level $\hat{\Pi}_i^\infty$ is approximately independent of $\lambda_i$, in agreement with the theoretical prediction $\tilde{\Pi}_{ii}^{\infty} \approx \tfrac12 \eta \gamma$.

\begin{figure}[t]
    \centering
    \includegraphics[width=0.95\textwidth]{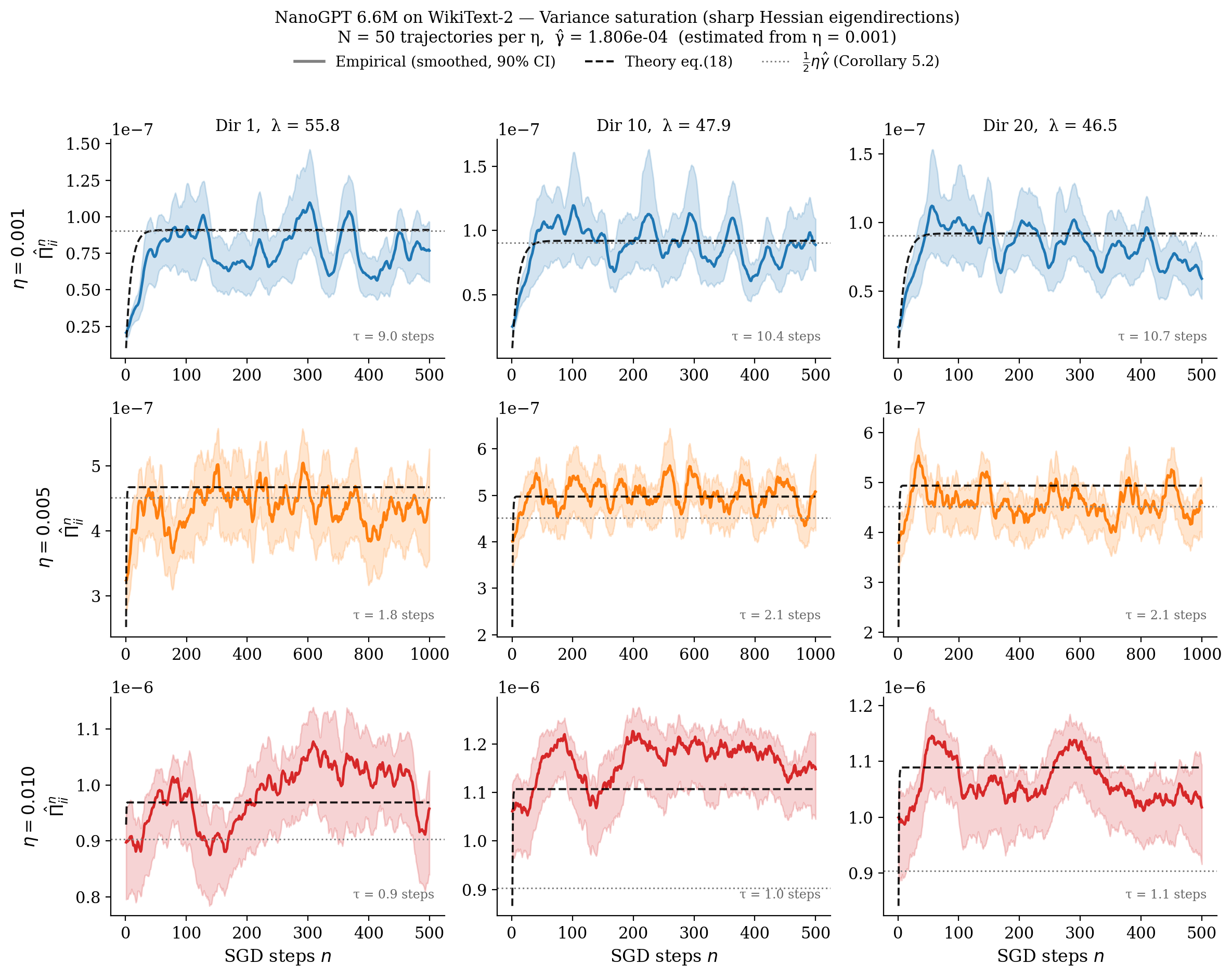}
    \caption{NanoGPT 6.6M on WikiText-2. Empirical variance $\hat{\Pi}^n_{ii}$ (solid, with $90\%$ bootstrap CI over $N=50$ trajectories) vs.\ theoretical prediction of eq.\ref{key} (dashed) for three sharp eigendirections and three learning rates. The dotted line shows the Corollary\ref{cor:regimes} approximation $\frac{1}{2}\eta\hat{\gamma}$. The theoretical time constant $\tau_i = (2\eta\lambda_i)^{-1}$ is annotated per panel. A single $\hat{\gamma} = 1.81 \times 10^{-4}$ is estimated from $\eta = 0.001$ and used to predict all other curves. Shaded bands show $90\%$ bootstrap confidence intervals obtained by resampling trajectories with replacement ($150$ resamples of $N=50$ from $50$), recomputing the variance estimate, and taking the $5$th and $95$th percentiles.}
    \label{fig:big_dynamics}
\end{figure}

At $\eta = 0.001$, the saturation curve exhibits a clearly visible exponential 
approach with $\tau \approx 9$--$11$ steps, and the empirical trajectory closely 
tracks the theoretical prediction of eq.\ref{key}. 
At $\eta = 0.005$ and $\eta = 0.010$, the saturation is nearly instantaneous 
($\tau \approx 2$ and $\tau \approx 1$ step respectively). 
The simple approximation $\tfrac{1}{2}\eta\hat{\gamma}$ (Corollary~\ref{cor:regimes}) 
deviates from the empirical plateau by up to $25\%$ at $\eta = 0.010$, while the 
full eq.\ref{key} — which retains the $\eta\,\mathbb{E}[\tilde{H}^2_{ii}]$ 
correction — remains within the empirical $90\%$ bootstrap CIs at all three 
learning rates, with a median relative error below $10\%$.

Figure~\ref{fig:big_eta} demonstrates that the mean saturation level scales linearly with $\eta$ across one order of magnitude, as predicted by Corollary\ref{cor:regimes}. The theoretical line $\frac{1}{2}\hat{\gamma}\eta$ with $\hat{\gamma}$ fixed from $\eta = 0.001$ achieves $R^2 = 0.90$ against the three empirical measurements, with the largest relative error at $\eta = 0.010$ ($25\%$) consistent with the growing importance of finite-step corrections at larger learning rates.

\begin{figure}[t]
    \centering
    \includegraphics[width=0.75\textwidth]{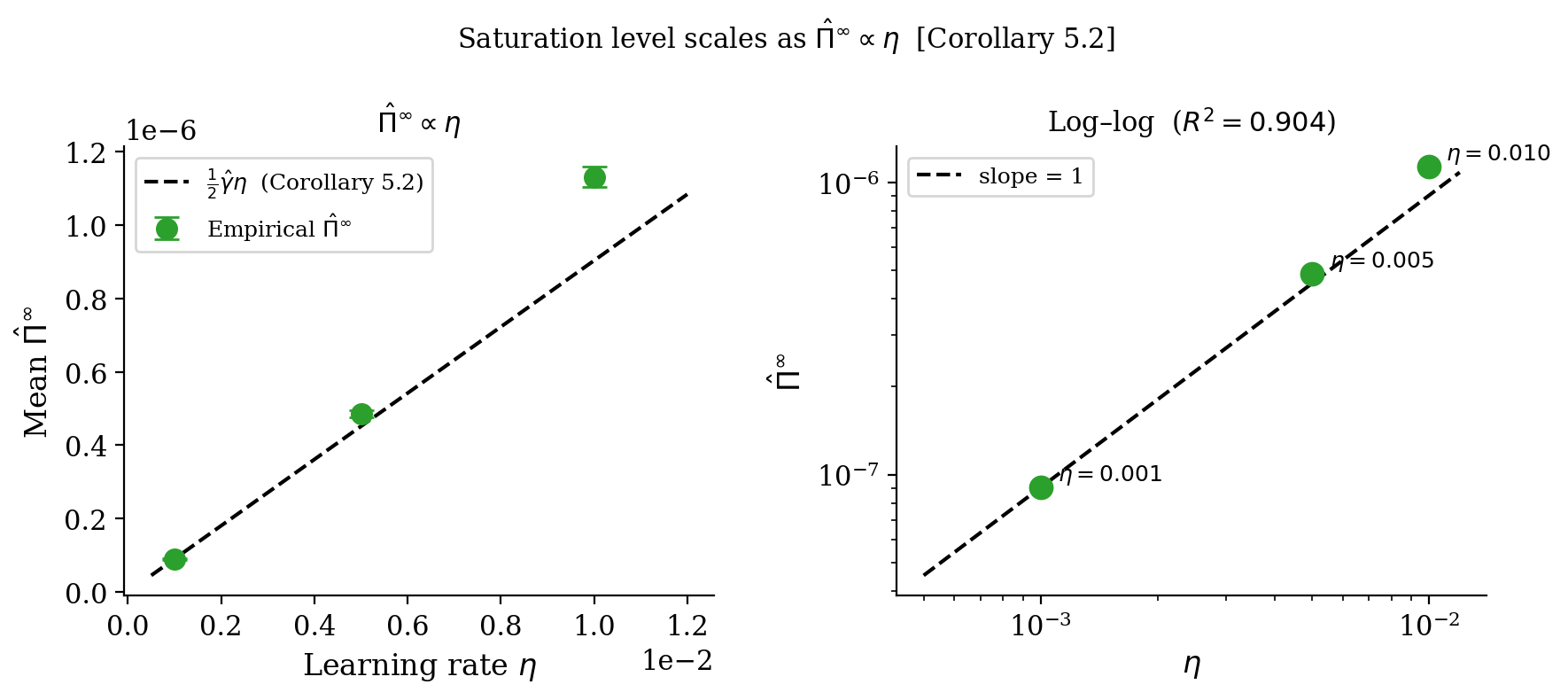}
    \caption{Mean saturation level $\overline{\hat{\Pi}^\infty}$ (averaged over $20$ sharp directions) as a function of learning rate $\eta$. Left: linear axes. Right: log--log axes with the theoretical slope-$1$ line. The single $\hat{\gamma}$ estimated from $\eta = 0.001$ predicts the other two points without refitting ($R^2 = 0.90$).}
    \label{fig:big_eta}
\end{figure}

\paragraph{Discrete SGD vs.\ Langevin approximation.}
We further compare the discrete prediction of eq.\ref{key} 
against the standard Langevin approximation 
$\Pi^{\mathrm{Lang}}_i = \eta d_i / (2\lambda_i - \eta\Gamma_{ii})$
(eq.~\eqref{eq:langevin_variance}), which differs from eq.\ref{key} 
only by the absence of the $-\eta\lambda_i^2$ term in the denominator. 
Using the same $\hat{\gamma}$ and the same noise ansatz $d_i = \hat{\gamma}\lambda_i$ 
for both predictions, Fig.~\ref{fig:big_level} shows that at $\eta = 0.001$ 
the two predictions agree within $3\%$, while at $\eta = 0.010$ the Langevin 
approximation systematically underpredicts the empirical plateau by ${\sim}23\%$ — 
an empirical confirmation at $6.6$M-parameter scale of the finite-step mismatch 
predicted by Proposition~4.2.

Taken together, these implementation details support the empirical validation presented in the main text and clarify how the quantities entering Eq.~\eqref{key} are estimated in practice at large scale.

\section{Additional Experiments}
\label{sec:appendix_experiments}

In this section, we provide additional experimental results supporting the main
claims of the paper. The experiments are designed to test four aspects of the
theory: (i) whether continuous-time Gaussian approximations reproduce discrete
SGD, (ii) whether the local covariance formula predicts the observed variance
profile in the Hessian eigenbasis, (iii) whether flat and sharp directions
exhibit distinct dynamical regimes, and (iv) whether the stochastic component
of SGD can be causally controlled. Unless stated otherwise, all experiments use
fixed random seeds, save the full configuration and raw intermediate artifacts,
and generate figures directly from saved outputs.

For empirical comparison, we construct discrete-time Gaussian surrogates driven
by the full gradient and a minibatch-noise estimate. Importantly, the stochastic
increment in these surrogates is scaled as $O(\eta)$, matching the scale of the
discrete SGD update, rather than adopting a literal continuous-time Brownian
discretization with externally specified noise. As a result, these baselines
should be interpreted as Gaussian approximations to SGD rather than exact
realizations of a Langevin diffusion.

\subsection{Quantitative Validation of the Discrete Theory}

We next validate the local covariance prediction of Eq.~\eqref{key}. Starting from a trained reference point, we compute the
full Hessian of the empirical loss and rotate trajectories into the eigenbasis
of the mean Hessian. The theoretical prediction is computed from independently
estimated gradient-noise statistics and Hessian eigenvalues; it is not obtained
by fitting the observed trajectory variance.

We perform this validation in two settings: the MLP-386 model on MNIST and the
NanoGPT character-level language model on Shakespeare. In both cases, the
reference point is obtained by training the model before the local Hessian and
trajectory analysis are performed. This ensures that the covariance comparison
is carried out in the same local regime as the one used in the theoretical
derivation.

For the MLP experiment, the predicted and measured variance profiles show strong
structural agreement across Hessian eigendirections. In the full run, the
log-correlation between predicted and measured variances is high, and held-out
checks show that the agreement is not explained by a single post-hoc rescaling.
Notably, the agreement is structural rather than merely scalar: the relative
ordering of variances across eigendirections is preserved.

We repeat the same validation for the NanoGPT model using the Shakespeare
character-level task. The model has 960 parameters, allowing exact dense Hessian
computation. The NanoGPT experiment follows the same protocol shape as the MLP
experiment: the reference point is trained inside the experimental pipeline,
the Hessian is computed at that point, and an ensemble of SGD trajectories is
launched from the same initialization. The predicted and measured variance profiles show strong structural agreement, confirming that the covariance prediction is not specific to a single architecture or modality.

We also estimate the proportionality constant $\gamma$ across multiple sampler
seeds and find it to be stable. Finally, we directly check the Hessian-noise
structure in the mean-Hessian eigenbasis: the diagonal mass dominates the
off-diagonal mass, supporting the approximate diagonality assumption used in
the derivation.

\begin{figure}[h]
    \centering
    \includegraphics[width=0.95\linewidth]{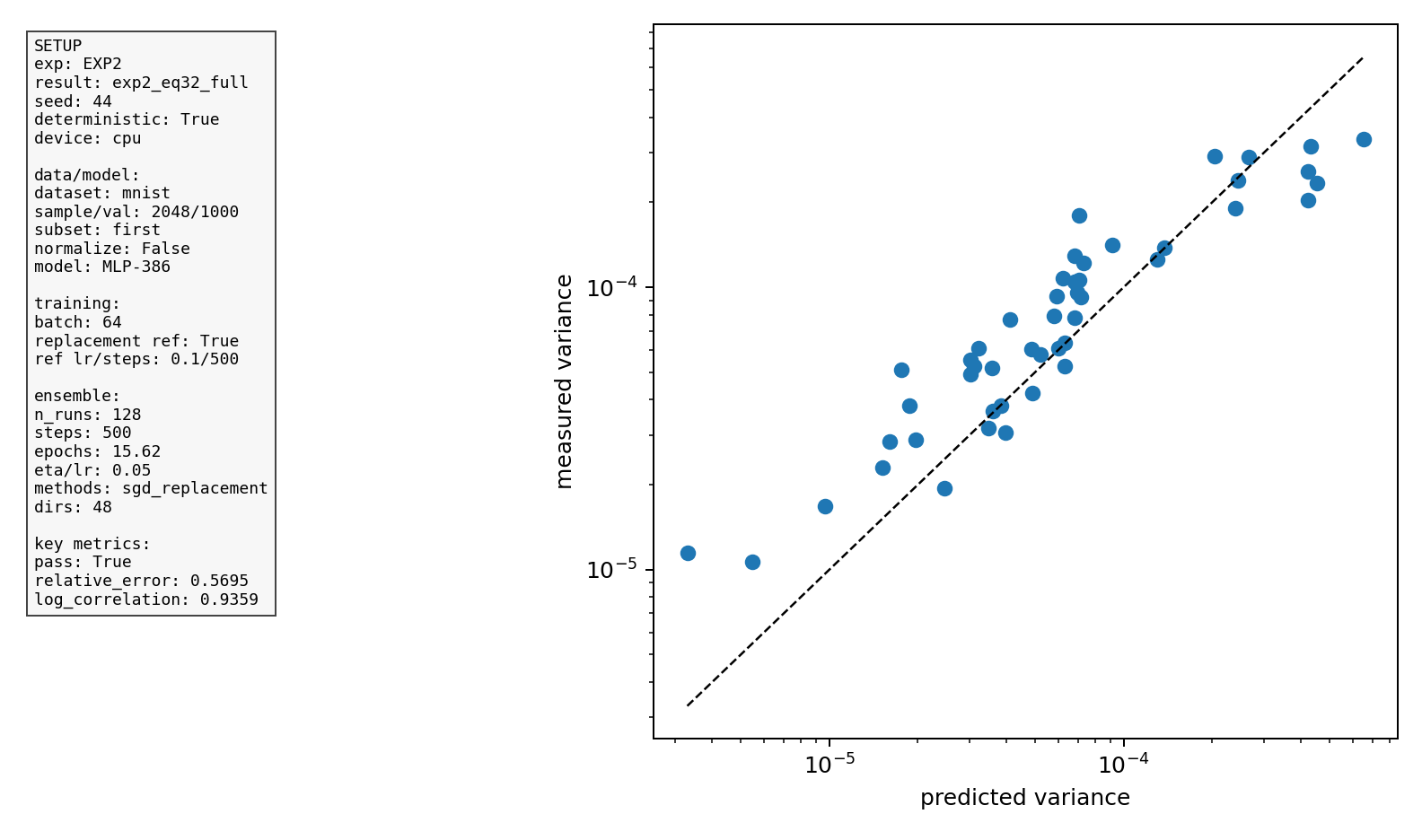}
    \caption{Quantitative validation of the discrete theory on MLP-386/MNIST. Predicted and measured variances across eigendirections show strong structural agreement.}
    \label{fig:appendix_exp2_eq32_mlp}
\end{figure}

\begin{figure}[h]
    \centering
    \includegraphics[width=0.95\linewidth]{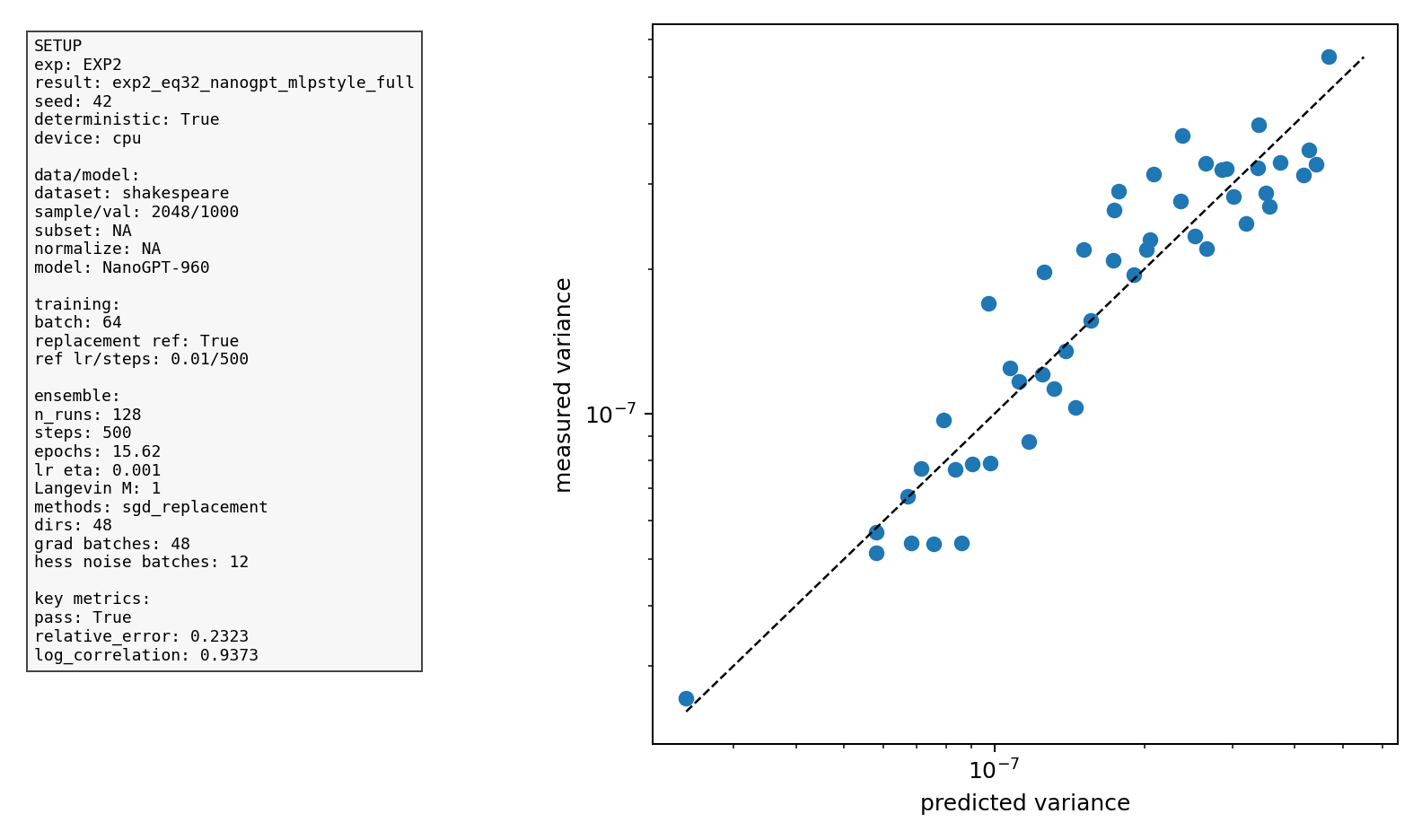}
    \caption{Quantitative validation of the discrete theory on NanoGPT/Shakespeare. The same covariance prediction remains structurally accurate for a small transformer language model.}
    \label{fig:appendix_exp2_eq32_nanogpt}
\end{figure}

\subsection{Sampling Effects}

Our theoretical master equation assumes sampling with replacement, ensuring independence of minibatches. In contrast, standard epoch-based training samples without replacement, introducing correlations between updates.

\paragraph{Remark (sampling and stochastic modeling).}
The sampling rule is therefore part of the probabilistic model, not only an
implementation detail. Langevin-type approximations usually rely on independent
minibatch noise, which is naturally matched by with-replacement sampling
\cite{li2017stochastic, mandt2017stochastic}. In contrast, reshuffling without
replacement introduces correlations between updates and is known to modify the
noise structure of SGD \cite{gurbuzbalaban2019random, haochen2020random}.

To test the size of this effect in our setting, we start from the same
MNIST/MLP-386 reference point and run trajectory ensembles under the two
sampling schemes, keeping the optimizer, batch size, learning rate, and number
of steps fixed. The trajectories are compared in the same projected coordinate
system using ensemble variance, mean-path displacement, and the projected mean
trajectory.

We observe that sampling without replacement significantly suppresses the ensemble variance and introduces structured temporal correlations. This highlights that the sampling rule is part of the stochastic model and affects the validity of Langevin-type approximations.

\begin{figure}[h]
    \centering
    \includegraphics[width=0.95\linewidth]{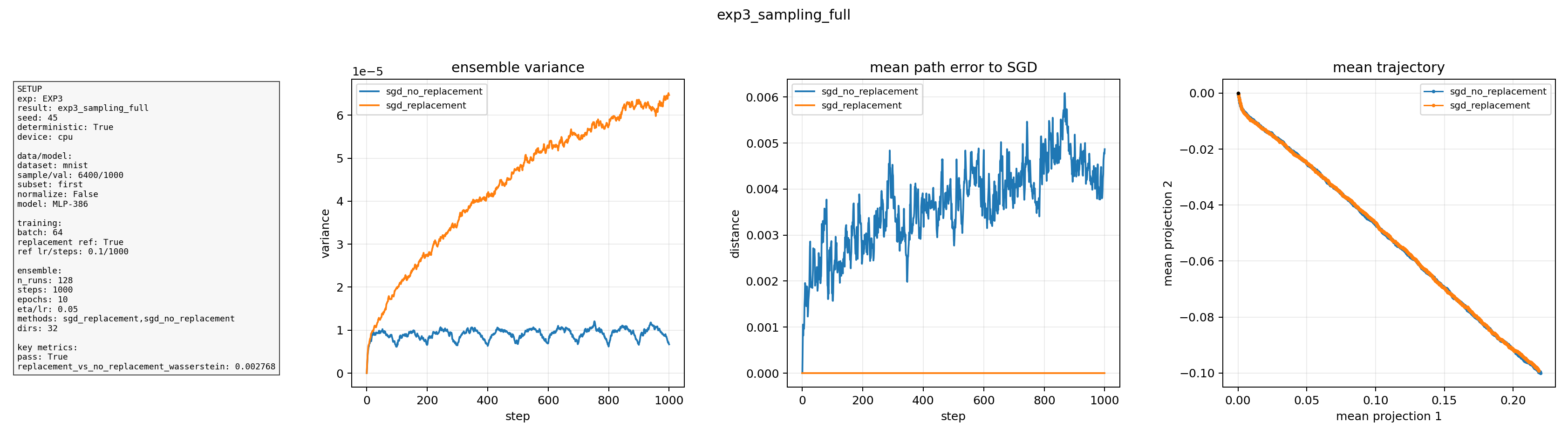}
    \caption{Effect of sampling strategy on SGD dynamics. Both ensembles start
    from the same reference point and use the same optimizer hyperparameters;
    only the minibatch sampling rule is changed. Left: mean variance over the
    projected directions. Middle: distance of each mean path to the
    with-replacement SGD mean path. Right: first two coordinates of the
    projected mean trajectory. Sampling without replacement strongly suppresses
    the projected ensemble variance, while the mean trajectory remains close to
    the with-replacement trajectory.}
    \label{fig:appendix_exp3_sampling}
\end{figure}

\section{Resources}

All experiments were intentionally conducted on small-scale models and synthetic systems in order to make exact second-order diagnostics feasible and fully reproducible.

For the MNIST experiments, we use an MLP with $386$ trainable parameters. This allows exact dense Hessian computation by second-order automatic differentiation. A full Hessian has size $386\times 386$, which is small enough to store and diagonalize directly on CPU.

For the NanoGPT experiments, the model has approximately $960$ trainable parameters. The corresponding dense Hessian has size $960\times 960$. These experiments remain tractable, but dense Hessian computation is already the dominant cost.

The additional reproducibility experiments in the exp6 suite were run on a local CPU machine with $12$ CPU cores and $16$GB RAM. CUDA was not used. The recorded full selected experiment pack completed in approximately one minute on this machine. Individual full runs ranged from below one second for small synthetic diagnostics to roughly $20$ seconds for the largest ensemble simulations. The largest observed process RSS in the recorded runs was below $0.5$GB, although this is only an end-of-run RSS snapshot rather than a true peak-memory measurement.

Trajectory-based experiments can require substantially more storage when run at the larger settings described in the paper. For example, storing $1000$ MLP trajectories of length $1111$ with $386$ parameters requires on the order of gigabytes. Similarly, storing many dense Hessians along a trajectory scales as $O(Td^2)$, where $T$ is the number of Hessian evaluations and $d$ is the number of parameters.

All reported figures are generated automatically from saved intermediate artifacts, including raw trajectories, covariance statistics, Hessian spectra, and metric files. No manual figure editing or spreadsheet processing is required.

These computational constraints are the reason the exact Hessian experiments are restricted to compact models. Extending the same diagnostics to larger modern architectures would require Hessian-vector products, Lanczos methods, or other approximate second-order techniques.

\end{appendixpart}
\end{document}